\documentclass[10pt,twocolumn,letterpaper]{article}

\usepackage[accsupp]{axessibility}  % Improves PDF readability for those with disabilities.

\usepackage{iccv}
\usepackage{times}
\usepackage{epsfig}
\usepackage{graphicx}
\usepackage{amsmath}
\usepackage{amssymb}

\usepackage[pagebackref=true,breaklinks=true,letterpaper=true,colorlinks,bookmarks=false]{hyperref}
\usepackage{booktabs}

\usepackage{makecell}
\usepackage{boldline}
\usepackage{algorithm, algorithmic}
\usepackage{multirow}  
\usepackage{enumitem}
\usepackage{breqn}
\usepackage{cleveref} % for cref
\usepackage{wrapfig}
\usepackage{subcaption} % for subfigure

\usepackage[table, x11names]{xcolor}
\definecolor{Gray}{gray}{0.93} 

\crefname{section}{Sec.}{Secs.}
\Crefname{section}{Section}{Sections}
\Crefname{table}{Table}{Tables}
\crefname{table}{Tab.}{Tabs.}
\crefname{figure}{Fig.}{Figs.}
\crefname{equation}{Eq.}{Eqs.}

\renewcommand{\etal}{\textit{et al.\ }}

\iccvfinalcopy % *** Uncomment this line for the final submission

\def\iccvPaperID{10766} % *** Enter the ICCV Paper ID here

\ificcvfinal\fi

\begin{document}

%%%%%%%%% TITLE
\title{DomainAdaptor: A Novel Approach to Test-time Adaptation}

\author{
Jian Zhang$^{1,2}$ \;\; Lei Qi$^{3,*}$ \;\; Yinghuan Shi$^{1,2,*}$ \;\; Yang Gao$^{1,2}$\\
$^1$~State Key Laboratory for Novel Software Technology, Nanjing University \\
$^2$~National Institute of Healthcare Data Science, Nanjing University \\
$^3$~School of Computer Science and Engineering, Southeast University \\
{\tt\small zhangjian7369@smail.nju.edu.cn, qilei@seu.edu.cn, syh@nju.edu.cn, gaoy@nju.edu.cn}
}

\maketitle
% Remove page # from the first page of camera-ready.
\ificcvfinal\thispagestyle{empty}\fi

\renewcommand{\thefootnote}{\fnsymbol{footnote}}
\footnotetext[1]{Corresponding authors: Yinghuan Shi and Lei Qi.
}

\begin{abstract}
  % To deal with the domain shift between training and test samples, current methods mainly focus on learning generalizable features during training.  However, these methods cannot capture the specificity of unseen samples during the test, and recent methods dealing with this problem cannot be directly applied to the trained model at test time. 
  To deal with the domain shift between training and test samples, current methods have primarily focused on learning generalizable features during training and ignore the specificity of unseen samples that are also critical during the test.
  In this paper, we investigate a more challenging task that aims to adapt a trained CNN model to unseen domains during the test. To maximumly mine the information in the test data, we propose a unified method called DomainAdaptor for the test-time adaptation, which consists of an AdaMixBN module and a Generalized Entropy Minimization (GEM) loss. Specifically, AdaMixBN addresses the domain shift by adaptively fusing training and test statistics in the normalization layer via a dynamic mixture coefficient {and a statistic transformation operation}. To further enhance the adaptation ability of AdaMixBN, we design a GEM loss that extends the Entropy Minimization loss to better exploit the information in the test data. Extensive experiments show that DomainAdaptor consistently outperforms the state-of-the-art methods on four benchmarks. Furthermore, our method brings more remarkable improvement against existing methods on the few-data unseen domain. The code is available at \url{https://github.com/koncle/DomainAdaptor}.
\end{abstract}

\section{Introduction}

 % Learning deep models under the independent identical distribution (IID) assumption has achieved tremendous success in various vision applications. However, this conventional assumption does not always hold when domain shift occurs, \ie, training source and test target data come from distinct domains.
 {To overcome the domain shift (\ie, training source and test target data come from distinct domains, for deep learning models)}, previous studies (\eg, unsupervised domain adaptation~\cite{na2021fixbi} or domain generalization~\cite{muandet2013domain}) have mainly focused on designing sophisticated models in the training stage. Despite their efforts, when a large domain gap exists in the test stage, they still inevitably suffer drastic performance degeneration.
Given that abundant information exists in the unlabeled unseen data during the test, which is neglected when solely considering generalization in the training phase, a more practical approach is to adapt a trained model to the unlabeled unseen data by incorporating this information during the test (\ie, test-time adaptation~\cite{wang2020tent}).

Incorporating both source data and unlabeled target data can improve adaptation and enable a model to handle unseen domains in real-world scenarios.  However, this approach is infeasible in practice due to the high computational cost of processing these data during the test. Besides, data privacy is paramount in many real-world scenarios where only model weights are accessible (\eg, clinical data~\cite{karani2021test,valanarasu2022fly} or commercial data~\cite{wang2020tent}). These restrictions remind us to consider a more practical problem: adapt a trained CNN model to an arbitrarily unseen domain in the test time without access to the source data, namely fully test-time adaptation~\cite{wang2020tent}, to expand the generalization ability of a trained model
  %. The key advantage of it is that it can be applied to any pre-trained model
  {without bearing retraining costs.}

% This argument has also been revealed in~\cite{zhang2021adaptive}. However, \cite{zhang2021adaptive} trains a model with a specifically designed method before adaptation, making its adaptation restricted, especially for some data privacy scenarios, \eg, clinical or financial data.
% Besides, for the models that have already been deployed and have limited generalization ability, retraining them for better generalization bears a large cost.
% 
%\bl{Note that we only perform single-shot adaptation (\ie, temporarily update model weights only once for the current batch without permanently changing the weights) instead of continuous adaptation (\ie, continuous updating the weights), considering the test data may come from different domains, which would degrade performance for continuous adaptation.}

\begin{figure}[t]
  \begin{center}
    \includegraphics[width=1.0\linewidth]{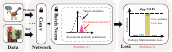}
  \end{center}
  \vspace{-10pt}
  \caption{{Illustration of the problem of (1) inaccurate statistics estimation in BN and (2) small loss produced by entropy minimization for the test-time adaptation methods.}}
  \label{fig_intro}
  \vspace{-5pt}
\end{figure}

Several methods~\cite{sun2020test,wang2020tent,boudiaf2022parameter} have been developed recently for fully test-time adaptation. One of the widely employed adaptation paradigm~\cite{wang2020tent, mummadi2021test,fleuret2021test,bateson2022test} to exploit the information of unseen samples is to finetune the Batch Normalization~\cite{ioffe2015batch} layers in a trained model with an unsupervised loss, which is both simple and computationally efficient.
{However, most of these methods only show their effectiveness on artificially corrupted datasets (\eg, CIFAR-10-C~\cite{hendrycks2019benchmarking}), which is different from the real-world scene with diverse cross-domain styles in the domain generalization task. For instance, when we adapt the model trained by photo data to the art painting data using existing methods, there still exist the following two issues, as shown in \cref{fig_intro}:}

{(1) The success of current test-time adaptation methods relies on an accurate estimation of the normalization statistics, which is hard to achieve by solely employing the test statistics obtained from the limited unseen data with a large domain gap. We argue that the source statistics can help the estimation, which is neglected by previous methods~\cite{wang2020tent}.}

{(2) During adaptation, previous unsupervised losses~\cite{carlucci2019domain} (\eg, Entropy Minimization (EM) loss) tend to bias the training procedure to the samples that have low confidence by producing large gradients and overlook the highly confident samples that also can help the adaptation procedure.}
% When the model is confident about the predictions of the unseen data, previous unsupervised losses (\eg, Entropy Minimization (EM) loss) are inefficient in providing sufficiently large gradients for effective adaptation. 
%{Finally, constrained by the drawbacks, previous fully test-time adaptation methods achieve satisfying performance only through online updating, which restricts its application to low-data scenarios.}
%{which obtains improved performance without a large number of testing data}.

Considering the above limitations of current methods, we present the DomainAdaptor, a novel approach for adapting a pre-trained CNN model to unseen domains, which comprises an AdaMixBN module and a Generalized Entropy Minimization (GEM) loss.
The AdaMixBN module overcomes inaccurate estimation of test statistics by combining training and test statistics and adapting the mixture coefficient based on the current batch. However, directly finetuning AdaMixBN may lead to performance degradation due to the weight mismatch problem caused by the combined source statistics after finetuning. To address this issue, we transform the source statistics into affine parameters in the normalization layers before finetuning. This not only maintains the effectiveness of AdaMixBN but also eliminates the negative impact of mismatched source statistics. Moreover, the traditional Entropy Minimization (EM) loss is not effective for finetuning AdaMixBN due to the sharp probability distribution predicted by the model for confident samples. Hence, we propose a GEM loss that emphasizes the role of temperature scaling in the traditional EM loss. GEM loss softens the probability distribution of each sample with temperature, generating large gradients for confident samples and encouraging further learning.
%Specifically, first, the discarded source statistics are important for stabilizing the statistics estimation~\cite{schneider2020improving}. Thus, to alleviate the inaccurate estimation of test statistics, we propose AdaMixBN, which combines training and test statistics for accurate statistics estimation and change the mixture coefficient adaptively according to the current batch.
%Besides, directly finetuning AdaMixBN may degrade performance significantly due to the weight mismatch problem caused by source statistics.  To mitigate this problem, we propose to remove the source statistics from the normalization procedure by transforming them into affine parameters in the normalization layers before finetuning, which could both keep the effectiveness of AdaMixBN and eliminate the negative effect of mismatched source statistics.
%Finally, traditional EM loss can not effectively finetune AdaMixBN with produced small gradients due to the sharp probability distribution predicted by the model for highly confident samples.  Therefore, to better finetune AdaMixBN, we propose the Generalized Entropy Minimization~(GEM) loss that emphasizes the role of temperature scaling in the traditional EM loss. By softening the probability distribution of each sample with temperature, GEM loss produces large gradients for highly confident samples and encourages further learning.

The contributions of our proposed DomainAdaptor can be summarized as follows:
\begin{itemize}[noitemsep,topsep=0pt]
  % \item We propose a novel DomainAdaptor that could effectively exploit the unlabeled batch data by adapting the trained model with a single finetuning.
  \item We propose AdaMixBN to adaptively mix the training and test stats. in the transformed normalization layer, which can trade off the training and test information.
        % \item A statistics transformation layer is proposed in AdaMixBN to prevent performance degeneration caused by statistics mismatch after finetuning.
  \item To better exploit unlabeled test samples, we propose the Generalized Entropy Minimization loss to effectively optimize the parameters of AdaMixBN.
  \item Our proposed method exhibits significant improvement over existing approaches on four benchmark datasets for domain generalization. %  on both online and few-data scenarios.
\end{itemize}

\section{Related Work}

% Since DG expects to generalize to unseen domains, learning invariant features across all domains~\cite{li2018deep,rahman2019correlation,zhao2020domain} by adversarial training~\cite{li2018domain,deng2020representation} or feature disentanglement~\cite{bousmalis2016domain, chen2021style} is employed to accomplish this goal. Except for only learning domain-invariant features, data augmentation-based methods utilize style transfer~\cite{yue2019domain,xu2020robust,zhou2020deep,li2020random}, statistics perturbations~\cite{zhou2021mixstyle,fan2021adversarially,shu2020prepare,jeon2021feature} or Fourier transformations~\cite{xu2021fourier} to augment diverse images to encourage the model to be robust to unseen styles. Besides, several methods design different training schemes~\cite{li2018learning,li2019episodic,li2019feature} or losses~\cite{carlucci2019domain} to regularize the model from overfitting.   

\textbf{Test time adaptation} (TTA)~\cite{sun2020test,fleuret2021test,wu2021domain,iwasawa2021test} is proposed to learn the test distribution by leveraging unlabeled test images, which provide hints about distribution information. Test-time training~\cite{sun2020test} employs a manually designed self-supervised learning task to learn the test distribution, which requires altering the training stage and finetuning all the layers. To mitigate this issue, Tent~\cite{wang2020tent} is proposed by only finetuning the batch normalization layers with an unsupervised entropy minimization loss. Following works try different unsupervised losses to help test time adaptation, such as consistency loss~\cite{fleuret2021test}, contrastive loss~\cite{liu2021ttt++} or log-likelihood ratio loss~\cite{mummadi2021test}. However, when applied to the test data with a large domain gap, these methods commonly fail due to the inaccurate estimation of statistics and produce small gradients. In contrast, our method can alleviate these issues and also succeed in the few-data scenarios.

\textbf{Domain generalization} (DG) has attracted significant attention recently for its ability to generalize to unseen domains by only learning from source domains~\cite{muandet2013domain,hendrycks2019benchmarking}.
To achieve the generalization ability, current methods primarily aim to learn invariant features across all domains~\cite{zhao2020domain,deng2020representation,chen2021style}, augment data~\cite{li2020random,zhou2021mixstyle,zhou2022generalizable,fan2021adversarially,xu2021fourier,guo2023aloft} to learn diverse features or regularize network with training schemes or losses~\cite{li2018learning,li2019episodic,li2019feature,carlucci2019domain,zhang2022mvdg}.
While these methods only consider the training stage, several methods alter model behaviors according to the test samples for better adaptation to the unseen domains. Instance Normalization~\cite{ulyanov2016instance} and AdaBN~\cite{li2018adaptive,zhang2022generalizable} are simple but effective modules that utilize test statistics to perform normalization. Du~\etal \cite{du2020metanorm} generates accurate statistics for each test sample with a trained statistics prediction network. In addition to normalization, Pandey~\etal \cite{pandey2021generalization} train a  generative network to generate the nearest neighbor for each test sample in the source latent space. ARM~\cite{zhang2021adaptive} applies a meta-learning training scheme to extract batch-specific features during training and test for adaptation. Despite their efforts to adapt the model at the test stage, they require modifying the training stage and cannot be applied to an already trained model. In contrast, we propose our DomainAdaptor, which can be employed in any trained model, making it more practical in the real world. 

\textbf{Temperature scaling} has been studied in different fields. In knowledge distillation~\cite{hinton2015distilling,lin2022knowledge,wu2022contextual,zhao2022decoupled}, it is used to soften the probability distribution over classes, providing additional class relationship information for the student model. In confidence calibration~\cite{guo2017calibration,ding2021local,tomani2021post,joy2022sample}, the temperature is finetuned as a parameter to ensure the predicted class confidence accurately reflects the likelihood of its ground truth correctness. In self-supervised learning~\cite{wang2021understanding,zhang2022dual,kukleva2023temperature}, the temperature is used in contrastive loss to penalize hard negative samples. Differently, in this work, we employ temperature to encourage effective learning by fully exploiting unlabeled test data.

% The normalization process ensures that the feature distribution is located in a reasonable range that the model can handle, which depends on the accurate estimation of statistics.  However, after finetuning, the original source statistics are not suitable for normalization since it is  

\begin{figure*}[t]
  \begin{center}
    \includegraphics[width=1.0\linewidth]{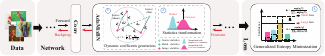}
  \end{center}
  \vspace{-5pt}
  \caption{\textbf{Method Overview}. With an unlabeled test batch, \textbf{(1)} our proposed AdaMixBN first obtains the dynamic mixture coefficient $\alpha$ with \cref{eq_alpha}. \textbf{(2)} Then, the affine parameters ($\gamma$,  $\beta$) are transformed with \cref{eq_affine} to avoid negative side-effect of mixed statistics. \textbf{(3)} Finally, we finetune the transformed affine parameters with Generalized Entropy Minimization (GEM) loss to fully exploit batch information. The labels of the same batch are predicted with the finetuned network.}
  \label{fig_framework}
  \vspace{-5pt}
\end{figure*}

\section{Our Method}

Let $\mathcal{X}, \mathcal{Y}$ denote the set of images and their corresponding label.  In the test stage, we are presented with a trained model $f(\cdot|\theta): \mathcal{X} \rightarrow \mathcal{Y}$ parametrized by $\theta\in \Theta$. At each timestep $t$, a batch of data points  $\{(x_1^t, y_1^t), \dots, (x_N^t, y_N^t)\}$ is sampled from the unseen joint distribution $\mathcal{P}_{\mathcal{X}\mathcal{Y}}$ and only the unlabeled batch of images $x^t=\{x_1^t, \dots, x_N^t\ \}$ are observed, where $N$ is the batch size. For simplicity, we drop the time $t$ in the following paragraphs.  Our goal is to predict the labels of the batch correctly by adapting the parameters $\theta$ of the trained model to the test distribution with only unlabeled data. To achieve this goal, we propose AdaMixBN with statistics transformation operation and Generalized Entropy Minimization loss to effectively exploit the information inside the unlabeled data. The whole framework is illustrated in \cref{fig_framework}.

\subsection{AdaMixBN}

\subsubsection{Dynamic coefficient generation}
\label{sec_alpha}
% Since the model is trained with source data, the Batch Normalization layers accumulate the source statistics to perform normalization. If these statistics are different from that of the unseen data, it would cause statistics mismatch and domain shift. Thus, to alleviate this issue, we can normalize the unseen batch with its own statistics~\cite{li2016revisiting}: 
% $$
%   \hat{{x}} = \frac{{x}-\mu_t}{\sqrt{\sigma_t^2+\epsilon}} \gamma + \beta,
% $$  

% Although it could reduce the domain shift, the inaccurate estimation of statistics leads to poor adaptation performance. Since the feature map in the high layers contains less domain-specific information~\cite{zhou2021mixstyle} and is more \textit{transferable} across domains (as shown in \cref{fig_alpha} in Experiments), the source statistics in the trained model ignored by Tent~\cite{wang2020tent} could also be utilized to help estimate a more accurate test statistics. 

% Therefore, to alleviate the abovementioned problem, we propose AdaMixBN, which utilizes source statistics to supplement the inaccurate test statistics. It mixes source statistics $(\mu_s, \sigma_s^2)$ and test statistics $(\mu_t, \sigma_t^2)$  with a dynamic mixture coefficient  $\alpha$:
% Therefore, AdaMixBN   It is inspired by \cite{schneider2020improving} that infers the statistics by combining the source  $(\mu_s, \sigma_s)$ and batch  $(\mu_t, \sigma_t)$statistics :
% \begin{align*}
%   \hat{\mu} =\alpha \mu_s + (1-\alpha) \mu_t, \ \ \ \  \hat{\sigma}^2  =\alpha \sigma_s^2 + (1-\alpha) \sigma_t^2.
% \end{align*}

Given a batch of data, despite the lack of label information, the statistics (\ie, mean and variance) of the extracted feature within the unlabeled data can also provide clues about the underlying distribution of the data, which can be utilized to help the model adapt to the corresponding domain. Since the difference between the source statistics accumulated during training and batch statistics during the test causes domain shift in the Batch Normalization layers~\cite{nam2018batch,fleuret2021test}, previous methods (\eg, Tent~\cite{wang2020tent}) only employ the test batch statistics for normalization and drop the source statistics, which is inaccurate for the real-world data and causes performance degradation. Considering that higher layers with high-level information are more transferable than lower layers (the experiment is conducted in \cref{sec_further}), the source statistics are also useful for adaptation. Therefore, we propose AdaMixBN that can dynamically fuse both statistics to obtain a more accurate statistics estimation during the test:
\vspace{-10pt}
\begin{align}
  \hat{x} = \frac{x-(\alpha\mu_s+(1-\alpha)\mu_t)}{\sqrt{(\alpha\sigma_s^2+(1-\alpha)\sigma_t^2)}}\gamma+\beta,
  \label{eq_adamixbn}
\end{align}
where $(\mu_s, \sigma_s^2)$ and $(\mu_t, \sigma_t^2)$ are the statistics estimated from the source and batch data, and $(\gamma, \beta)$ are the affine parameters in BN. The layer index is omitted for simplicity.

\begin{table}[t]
  \center
  \caption{The average distances between the single test image,  test batch, and the source statistics in the network.}
  % \vspace{-5pt} 
  \renewcommand\arraystretch{0.95}
  \resizebox{0.95\columnwidth}{!}{
    \begin{tabular}{ l  | c  c c c | c }
      \toprule
      \textbf{Statistics Distance} & \textbf{Art} & \textbf{Cartoon} & \textbf{Photo} & \textbf{Sketch} & \textbf{Avg.} \\
      \midrule
      img2batch (Min)              & 0.016        & 0.013            & 0.013          & 0.000           & 0.011         \\
      img2batch (Max)              & 3.334        & 5.418            & 2.772          & 0.061           & 2.896         \\
      % img2test (Mean) & 0.669      & 0.472      & 0.402      & 0.006      \\
      src2batch                    & 1.031        & 0.269            & 1.148          & 3.251           & 1.425         \\
      \bottomrule
    \end{tabular}
  }
  \vspace{-5pt}
  \label{tab_dist}
\end{table}

Instead of simply employing a manually defined fixed mixture coefficient $\alpha$ which cannot handle the varying unseen domains, we design a novel generation module to dynamically mix the training and test statistics according to the Euclidean distance $d=\|\mu_1-\mu_2\|_2+\|\sigma_1-\sigma_2\|_2$ of the statistics between source and test statistics:
\begin{align}
  \alpha = 1-\frac{1}{N}\sum_{i}\frac{d_{st}}{d_{t}^i+d^i_{s}}.
  \label{eq_alpha}
\end{align}
There are three distances in the formula: $d_{st}$ is the distance between source statistics and test statistics; $d^i_{s}$ and $d^i_{t}$ are the distances of the statistics of a single image $x_i$ to the source and batch statistics, respectively. Although a single distance $d_{st}$ can be employed to calculate $\alpha$, a fixed threshold is required to determine its relationship to the value of $\alpha$. For instance, $\alpha$ can be obtained as $\alpha=\textrm{sigmoid}(d_{st} - \textrm{threshold})$, where the threshold needs to be manually set for each batch, which is impossible.  We observe that the statistics of the images in the batch should not be far away from the overall batch statistics.
If the source-to-batch distance $d_{st}$ is smaller than the image-to-batch distance $d_t^i$, this implies that the source domain is similar to the test domain, and a large $\alpha$ can be used to incorporate more source statistics, as shown in \cref{fig_framework}.
{For example, in the above \cref{tab_dist}, the larger source-to-batch distance than the image-to-batch distance indicates a large domain gap for the Sketch domain. To reduce this domain gap, more test statistics should be incorporated (\ie, a small $\alpha$).}
Therefore, instead of manually designing $\alpha$ or adopting a threshold,  we propose to employ the single image statistics as a proxy to dynamically measure the distance. We introduce the image-level distances $d_s^i$ and $d_t^i$ as a relative distance measure that adapts to different batches and layers.
When the source and target statistics differ significantly, $d_{st}$ and $d_{s}^i$ would be large while $d_t^i$ keeps the same. Then the ratio would be large and result in a small $\alpha$, and vice versa. We average the ratio to obtain the final $\alpha$.

% \re{Using more test statistics actually leads to performance degradation, as shown in AdaMixBN (Reverse) in \cref{compare}.
% The reason is as follows:
% 1) the source statistics, accumulated with all source data, are more reliable than the test statistics calculated with limited samples;
% 2) incorporating more source statistics is only viable when the source and test statistics exhibit similarity; otherwise, excessive incorporation can cause a large distributional shift.
% {Our principle to choose the $\alpha$ also follows this intuition that we require a measurement that returns a high ratio when the test statistics is close to the source statistics.
% Considering the diverse statistics of the individual image,
% }}

\begin{table}[t]
  \center
  \caption{The test accuracy (\%) before/after fientuning (FT) with different statistics. T is statistics transformation.}
  \resizebox{0.8\columnwidth}{!}{
    \begin{tabular}{ l  | c  c c | c  }
      \toprule
                & \textbf{Test} & \textbf{Mixed}      & \textbf{Source}     & \textbf{Mixed w/ T} \\
      \midrule
      Before FT & 80.53         & {83.43}             & 79.94               & 83.43               \\
      After FT  & 81.96         & {65.01$\downarrow$} & {17.36$\downarrow$} & 85.04               \\
      % \hline
      % Distribution shift & 0.112           & 0.001           & 0.022        & 0.001             \\
      \bottomrule
    \end{tabular}
    \label{finetune_changes}
  }
  \vspace{-5pt}
  \label{tab_ft}
\end{table}

\subsubsection{Statistics transformation}
\label{sec_transformation}

% \re{To further boost the performance, we try to update the source-trained affine parameters during the test stage. However, its performance obviously decreases (\ie, 65.01) due to the explicitly mixed source statistics, as shown in this table. 
% The reason is that finetuning the affine parameters in BN alters the feature map and causes a discrepancy between source and feature statistics. However, the performance of using dynamic test statistics does not degrade. Therefore, \textit{to both keep the better performance of mixed statistics while eliminating the side-effect of source statistics}, we propose to incorporate the source statistics \underline{in an implicit way} by transforming the normalization formulation of AdaMixBN to match the formulation that only incorporates the test stats., which is our intuition.}   

% To eliminate the fixed source statistics problem, we propose to move the source statistics into the affine parameters \textit{before finetuning}, such that the normalization process does not incorporate the source statistic as shown in the bottom row of \cref{fig_transform}. 

Although AdaMixBN can improve the test-time performance, directly finetuning the network with AdaMixBN leads to significant degradation in performance.
As shown in \cref{tab_ft}, when fintuning with only test statistics, the performance is improved, while when incorporating more source statistics in the normalization process, the performance degrades significantly. To the extreme, if we only employ source statistics without using test statistics, the performance cannot even defeat the random guess.

\begin{figure}[t]
  \begin{center}
    \includegraphics[width=1.0\linewidth]{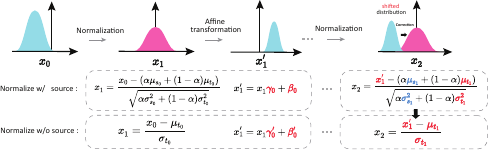}
  \end{center}
  \vspace{-10pt}
  \caption{ The normalization process after finetuning with only test (w/o source) or mixed statistics (w/ source).}
  %After finetuning, the affine parameters are changed and produce a changed feature map $x_1'$ (marked as red), while source statistics are fixed (marked as blue). Using the statistics mixed with source statistics causes a distribution shift (top right figure), and the statistics transformation addresses it by transforming the source statistics into affine parameters \textit{before finetuning}. }
  \label{fig_transform}
  \vspace{-5pt}
\end{figure}

This problem arises due to the weight mismatch between the finetuned parameters (\ie, $\gamma, \beta$ in all BN layers) and source statistics, as illustrated in the middle row of \cref{fig_transform}.
After finetuning, BN weights (\eg, $\gamma_0$ and $\beta_0$ in \cref{fig_transform}) are changed, which outputs the feature map $x_1'$ with the shifted distribution. However, the source statistics (\eg, $\mu_{s1}$ and $\sigma_{s1}^2$ in \cref{fig_transform}) are fixed across the finetuning process. Then if we continue to utilize the source statistics that are only suitable for the original distribution to normalize feature map $x_1'$ that has shifted distribution, the performance inevitably degrades. The more source statistics involved, the more serious the degradation. However, the performance of using dynamic test statistics does not suffer from this degradation.
Therefore, \textit{to both keep the better performance of mixed statistics while eliminating the side-effect of source statistics}, we propose to incorporate the source statistics \textit{in an implicit way} by transforming the normalization formulation of AdaMixBN to match the formulation that only incorporates the test statistics.
To achieve this goal, we can transform the normalization process of AdaMixBN in \cref{eq_adamixbn} to be independent of the source statistics (\ie, $\mu_s$ and $\sigma_s$). With several simple deductions, we obtain the following transformation (details are in Supplementary Material):
\begin{align}
  \hat{x} & =\frac{x-\mu_t}{\sigma_t} \gamma' + \beta',                                     \\
  \gamma' & = \frac{\sigma_t}{\sqrt{\alpha \sigma_s^2+(1-\alpha)\sigma^2_t}}\gamma + \beta, \\
  \beta'  & =  \frac{\alpha(\mu_t-\mu_s)}{\sigma_t} \gamma' + \beta.
  \label{eq_affine}
\end{align}
This transformation can be viewed as re-initializing $\gamma$ and $\beta$ using the source and test statistics. In this way, the normalization process only depends on the test statistics $\mu_t$ and $\sigma_t$, which can dynamically change according to the distribution of the current feature map. Note that, the operation is done after the batch has been fed into the network and \textit{before} the finetuning process. As a result, we finetune $\gamma'$ and $\beta'$ in each layer instead of the initial affine parameters. Furthermore, since we have altered the normalization process, the training and test mode of BN can not affect it.

\subsection{Generalized Entropy Minimization}
\label{sec_emt}

\subsubsection{GEM framework}

After feeding an unseen batch into the model and transforming its affine parameters for test-time adaptation with AdaMixBN, we can adapt the model with its predictions. However, we can not perform supervised finetuning as in few-shot learning~\cite{finn2017model} since the label is unavailable. Instead, we can consider using several unsupervised losses with human prior (\eg, rotation prediction task~\cite{sun2020test} or jigsaw task~\cite{carlucci2019domain,noroozi2016unsupervised}). Different from these losses that require manually defined labels, Entropy Minimization~(EM) loss is a simple but effective unsupervised loss that learning from data by encouraging the model to be confident~\cite{wang2020tent}:
\vspace{-2pt}
\begin{align}
  \mathcal{L}_{\mathrm{EM}} = -\sum_{i=1}^C p_i \log p_i, \ \  p_i=\frac{\exp(z_i/\tau_p)}{\sum_{k=1}^C\exp(z_k/ \tau_p)},
  \label{eq_em}
\end{align}
where $z_i$ is the logits predicted by the model, $C$ is the number of classes and $\tau_p=1$ is the temperature of the EM loss.

\begin{figure}[t]
  \begin{center}
    \includegraphics[width=0.98\linewidth]{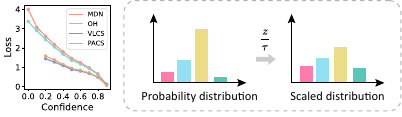}
  \end{center}
  \vspace{-10pt}
  \caption{Loss decreases when confidence increases (left). Probability changes after temperature scaling (right). }
  \label{fig_emt}
  \vspace{-5pt}
\end{figure}

% To encourage learning from the highly confident samples, SLR~\cite{mummadi2021test} utilizes a log-likelihood ratio loss to encourage the model to produce a high ratio between the highly confident classes and low-confident classes. However, it pays more attention to the highly confident samples and ignores the low-confident samples.
During the finetuning of a network with EM loss, highly confident samples that have a high probability of predicted classes, produce small contributions for weight updating as illustrated in the left figure of \cref{fig_emt}, resulting in little improvement on the trained model.
We observe that highly confident samples have a sharp logit distribution as demonstrated in the right figure in \cref{fig_emt}.
To reduce the sharp distribution, we propose to use temperature scaling for two reasons. First, it does not change the model's prediction, which can preserve the original semantic information. Second, for highly confident samples, we can obtain a larger loss by softening the logit distribution to enable fast learning.  Meanwhile, the smoothness effect of temperature scaling decays when a sample has lower confidence, and the model also can learn from the low-confident samples. To this end, we propose a novel Generalized Entropy  Minimization~(GEM) framework with temperature scaling to adapt the trained model to the test unlabeled data:
\begin{align}
  \mathcal{L}_{\mathrm{GEM}} = -\tau^2_q\sum_{i=1}^C p_i \log q_i,  q_i = \frac{\exp{(z_i/\tau_q)}}{\sum_{k=1}^C\exp{(z_k/\tau_q)}},
  \label{eq_gem}
\end{align}
where $\tau_q$ is the temperature for $q_i$, and we set $\tau_q\ge\tau_p\ge1$ to soften the  logits distribution. Note that we multiply GEM loss by $\tau_q^2$ to ensure normal gradient magnitudes since the temperature scaling downscales not only the logits to $1/\tau_q$, but also the gradients to $1/\tau_q^2$~\cite{hinton2015distilling}. Different from traditional EM loss that employs the same probability in \cref{eq_em}, we utilize two different probabilities as a general version of the cross-entropy loss, which imposes two different behaviors of GEM loss as analyzed in the following section.

\subsubsection{Gradient analysis of GEM and GEM variants}
\label{sec_grad}

For the logit $z_k$ of class $k$, the gradient from GEM loss is (details can be found in Supplementary Material):
% \begin{align}
%    \frac{\partial L_{GEM}}{\partial z_{k}} &= \notag -\tau_q^2[\frac{1}{\tau_q} \left(p_{k}-q_{k}\right) 
%    +\frac{p_{k}}{\tau_p} (\log q_{k}-\sum_{j} p_{j}  \log q_{j})].
%  \label{eq_grad}   
% \end{align}    
\begin{align}
   & \frac{\partial L_{\mathrm{GEM}}}{\partial z_{k}} = \notag -\tau_q^2[\frac{1}{\tau_q} \left(p_{k}-q_{k}\right) \\
   & \phantom{ \frac{\partial L_{\mathrm{GEM}}}{\partial z_{k}} =}
  +\frac{p_{k}}{\tau_p} (\log q_{k}-\sum_{j} p_{j}  \log q_{j})].
  \label{eq_grad}
\end{align}
There are two terms in the right hand. If the gradient of $p_i$ is stopped in \cref{eq_gem} (similar to the one-hot label in the supervised cross-entropy loss), only the first term in \cref{eq_grad} is left, which actually takes the same form as the knowledge distillation loss employed in the \textit{teacher-student} network~\cite {hinton2015distilling}, where $p_k$ can be viewed as the probability output from the teacher model and $q_k$ from the student model. Optimizing with the first term could encourage the model to output the distribution similar to $p$. Therefore, the first term produces a Self-Knowledge Distillation (SKD) gradient that learns from itself. By increasing the temperature of $q$, the discrepancy between $p_k$ and $q_k$ would be large, and a larger gradient is produced. Besides, if $\tau_q < \tau_p$, the distribution of $q$ is sharper than $p$. The model will learn from a softer distribution and perform poorer. Thus, the $\tau_p$ should always be greater or equal to $\tau_q$ for better performance. When $p_k=q_k$, the first term is zero, and only the second term is left in \cref{eq_grad}, which plays the same role in the original entropy minimization loss that sharpens the distribution.
Note that neither of these two terms can change the model's prediction when only a single sample is involved since they can only produce sharper logits distribution without changing its prediction. Instead, we can learn diverse features and improve the performance by \textit{learning from a batch}.

% the norm of logits increased after finetuning with GEM loss and produced highly confident predictions. Therefore, the magnitude of the logits' norm is also a measure of the model uncertainty. The more certain, the larger the logits' norm. 
% However, if we simply maximize the norm of the feature without changing its angle, the model cannot learn from the unlabeled data, which is verified in the Supplementary Material. 

% Furthermore, to adaptively change $\tau$ according to different logits distritbuions, we follow previous work~\cite{jia2020spherical} that uses the standard deviation of logits $\sigma_{z_i}$ to adaptively scale the logits $\tau=s\cdot\frac{1}{N}\sum_{i=1}^N\sigma_{z_i}$, where $s$ is a hyper-parameter to adjust the scaling strength.
Under the GEM framework, when $\tau_p=\tau_q=1$ , GEM loss is the same as the vanilla EM loss.
We notice that $p_i$ can be viewed as a weighting factor for GEM loss.  If $p_i$ is a one-hot tensor, only one term is left after summation. It is actually the same as the traditional cross-entropy loss that encourages the model to learn from a single class. But when $p_i$ becomes smoother (\ie, $\tau_p$ is larger), it encourages the model to learn from all classes. Thus, we propose several variants by setting different $\tau_p$ to produce different $p_i$.  For $\tau_q$,  we set it to $\tau_q=\tau_a=\frac{s}{N}\sum_{i=1}^N\sigma_{z_i}$, where $s$ is a hyper-parameter to adjust the scaling strength, which follows previous work~\cite{jia2020spherical} that uses the standard deviation of logits $\sigma_{z_i}$ as a dynamic temperature to scale the logits.

\textbf{GEM-T.} We set $\tau_p=\tau_a$. The first term in \cref{eq_grad} becomes $0$, and both probabilities are scaled smoother to produce a large gradient. In this way, since all classes in the distribution are scaled and no distribution of samples is too sharp, the model is easier to make different predictions.

\textbf{GEM-SKD.} We set $\tau_p=1$ and the gradient of $p_i$ is stopped. In this manner, we only employ the Self-Knowledge Distillation term in \cref{eq_grad} for learning, where $p_i$ is teacher's output. Different from GEM-T, which encourages learning from all classes, it prevents the model from changing too far away from its original prediction.

\textbf{GEM-Aug.} Being aware that $p_i$ can be viewed as a teacher's prediction in the GEM-SKD loss, the distillation process can be further improved if $p_i$ is more accurate. Thus, we propose to adopt Test Time Augmentation to logit $z_i$ in GEM-SKD to obtain a more accurate prediction of $p_i$. $p_i= \mathrm{softmax}\big(\frac{1}{m}\sum_{j=1}^{m} z^j_i\big),$ where $m$ is the number of augmented images for each sample.

Our DomainAdaptor equipped with these three different GEM losses is named DomainAdaptor-T, DomainAdaptor-SKD, and DomainAdaptor-Aug, respectively.

  {\textbf{Remark:}  1) Note that, with the above contributions that fully take advantage of unlabeled data by fusing source and target statistics and scaling the temperature, our method can adapt to the data with a single finetune step without permanently changing its weights. 2) Benefiting from it, DomainAdaptor offers a significant advantage over previous methods that require online updating of model weights, making it suitable for the few-data unseen domain. This will be demonstrated in the experiment section.}

\section{Experiments}

In the experiment, we first compare our method to previous methods and analyze the ablation study of each component. Next, we apply our method to several trained DG SOTA methods to validate its wide application. Finally, we further analyze the hyperparameter sensitivity of our proposed AdaMixBN and how GEM loss works. More experiments can be found in the Supplementary Material.

% Experiments on DomainBed?
% \textbf{DataSets}. 
We employ four domain generalization datasets to test the performance of adapting to unseen domains with a large distributional gap, including \textbf{PACS}~\cite{li2017deeper}, \textbf{VLCS}~\cite{torralba2011unbiased}, \textbf{OfficeHome} (OH)~\cite{venkateswara2017deep}, and \textbf{MiniDomainNet} (MDN)~\cite{zhou2021domain}, a  subset of DomainNet~\cite{peng2019moment}. PACS is composed of $9,991$ images with $7$ classes. VLCS contains $10,729$ images with $5$ classes. OfficeHome has $15,500$  images with $65$ classes. MiniDomainNet consists of $140,006$ images with $126$ classes. All of these datasets contain $4$ domains. PACS and MiniDomainNet have a large domain gap, while VLCS and OfficeHome have a relatively small domain gap. Following the DG training and test scheme, we leave one domain out as the test domain and the others as training domains to pre-train the network. More experimental details can be found in the Supplementary Material.

\subsection{Comparison to Previous Methods}

We compare our method to several test-time adaptation methods (\ie, Tent~\cite{wang2020tent}, SLR~\cite{mummadi2021test}, LAME~\cite{boudiaf2022parameter}, and ARM~\cite{zhang2021adaptive}) as shown in \cref{tab_sota}. {Note that the methods employing online updating are marked with (o) while the others update the parameter temporarily for the current batch.} In these methods, ARM requires altering the training stage and has its own trained network (denoted as ARM (Base)), while other methods utilize DeepAll as the trained network.   The performance of the ARM baseline is low (\eg, $74.49\%$ vs.\ $79.44\%$) because the weight space that the meta-learning training paradigm finds is not suitable for a specific domain, which needs to be finetuned in the test stage to achieve better performance. Besides, Tent, SLR, and ARM all adopt AdaBN~\cite{li2018adaptive} that employs test batch statistics for normalization and finetuning the network during adaptation. Differently, LAME does not perform AdaBN and finetune. Instead, it exploits the relationship between images in a batch to refine the predictions with iterative optimization.

\begin{table}[t]
  \begin{center}
    \renewcommand\arraystretch{1.05}
    \caption{Comparison to SOTA with Resnet-18 and Resnet-50 as backbone.  The best performance (\%) is marked as \textbf{bold}. Methods marked with (o) are updated online.} %methods denoted with * are equipped with AdaMixBN.}
    \resizebox{1\columnwidth}{!}{
      \begin{tabular}{ l  | c  c c c | c }
        \toprule
        % \multicolumn{6}{c}{\textbf{PACS}}                                                                                     \\
        % \midrule
                                             & \textbf{PACS}              & \textbf{VLCS}               & \textbf{OH}                 & \textbf{MDN}                & \textbf{Avg.}    \\
        \midrule
        \multicolumn{6}{c}{\textbf{\textit{Resnet-18}}}                                                                                                                                \\
        \midrule
        DeepAll                              & $ 79.44_{\pm0.44}$         & $ 75.77_{\pm0.29}$          & $ 64.61_{\pm0.18}$          & $ 65.12_{\pm0.11}$          & $71.24$          \\
        ARM(Base)~\cite{zhang2021adaptive}   & $ 74.49_{\pm3.86}$         & $ 72.10_{\pm0.51}$          & $ 63.33_{\pm1.22}$          & $ 64.97_{\pm0.33}$          & $68.72$          \\
        % \midrule
        AdaBN~\cite{li2018adaptive}          & $ 80.44_{\pm0.29}$         & $ 69.44_{\pm0.48}$          & $ 63.38_{\pm0.12}$          & $ 64.20_{\pm0.11}$          & $69.37$          \\
        ARM~\cite{zhang2021adaptive}         & $ 82.47_{\pm0.59}$         & $ 68.21_{\pm1.68}$          & $ 63.15_{\pm0.61}$          & $ 64.91_{\pm0.23}$          & $69.69$          \\
        % Norm    & $ 83.18_{\pm0.20}$           & $ 71.37_{\pm0.47}$          & $ 62.83_{\pm0.13}$          & $ 63.05_{\pm0.07}$                   \\
        SLR~\cite{mummadi2021test}           & $ 81.33_{\pm0.22}$         & $ 69.69_{\pm0.44}$          & $ 63.64_{\pm0.09}$          & $ 64.51_{\pm0.10}$          & $69.79$          \\
        {SLR (o)}~\cite{mummadi2021test}     & $\textbf{85.15}_{\pm0.35}$ & $74.13_{\pm0.58}$           & $64.71_{\pm0.12}$           & $43.69_{\pm0.84}$           & $66.92$          \\
        Tent~\cite{wang2020tent}             & $ 80.59_{\pm0.26}$         & $ 69.69_{\pm0.44}$          & $ 63.58_{\pm0.14}$          & $ 64.37_{\pm0.09}$          & $69.56$          \\
        {Tent (o)}~\cite{wang2020tent}       & ${83.56}_{\pm0.47}$        & ${73.37}_{\pm0.31}$         & ${64.54}_{\pm0.06}$         & $48.50_{\pm1.86}$           & $67.49$          \\
        LAME~\cite{boudiaf2022parameter}     & $ 80.28_{\pm0.33}$         & $ 75.59_{\pm0.96}$          & $ 63.16_{\pm0.28}$          & $64.42_{\pm0.28}$           & $70.86$          \\
        \midrule
        DomainAdaptor-T                      & $ {85.04}_{\pm0.23}$       & $ 77.54_{\pm0.14}$          & $ 65.39_{\pm0.19}$          & $ 66.39_{\pm0.08}$          & $73.59$          \\
        DomainAdaptor-SKD                    & $ 84.37_{\pm0.28}$         & $ 78.10_{\pm0.14}$          & $ 65.61_{\pm0.14}$          & $ 66.42_{\pm0.08}$          & $73.63$          \\
        DomainAdaptor-Aug                    & $ 84.93_{\pm0.19}$         & $ \textbf{78.50}_{\pm0.22}$ & $ \textbf{66.73}_{\pm0.25}$ & $ \textbf{68.23}_{\pm0.08}$ & $\textbf{74.60}$ \\
        % \midrule
        % LAME'     & 83.40                       & $1_{\pm0.59}$               & $ 64.14_{\pm0.29}$          & $ 65.68_{\pm0.08}$          \\
        % Norm'     & \textbf{$ 85.00_{\pm0.21}$} & $ 77.48_{\pm0.12}$          & $ 64.79_{\pm0.10}$          & $ 2.04_{\pm0.05}$           \\
        % SLR'      & $ 84.04_{\pm0.21}$          & $ 77.35_{\pm0.26}$          & $ 65.23_{\pm0.12}$          & $ 63.17_{\pm0.07}$          \\
        % Tent'     & $ 83.63_{\pm0.21}$          & $ 77.14_{\pm0.36}$          & \textbf{$ 65.46_{\pm0.20}$} & $ 66.20_{\pm0.10}$          \\
        \midrule
        \multicolumn{6}{c}{\textbf{\textit {Resnet-50}}}                                                                                                                               \\
        \midrule
        DeepAll                              & $84.37_{\pm0.42}$          & $ 76.96_{\pm0.54}$          & $ 70.87_{\pm0.16}$          & $ 71.49_{\pm0.10}$          & $75.92$          \\
        ARM (Base) ~\cite{zhang2021adaptive} & $70.37_{\pm7.19}$          & $77.49_{\pm0.71}$           & $49.01_{\pm3.01}$           & $71.51_{\pm0.11}$           & $67.08$          \\
        AdaBN ~\cite{li2018adaptive}         & $85.29_{\pm0.43}$          & $ 70.95_{\pm0.61}$          & $69.37_{\pm0.24}$           & $70.21_{\pm0.08}$           & $73.96$          \\
        ARM ~\cite{zhang2021adaptive}        & $86.10_{\pm0.74}$          & $78.28_{\pm0.71}$           & $66.66_{\pm1.00}$           & $70.85_{\pm0.10}$           & $75.50$          \\
        SLR~\cite{mummadi2021test}           & $ 86.06_{\pm0.47}$         & $ 71.75_{\pm0.62}$          & $ 69.61_{\pm0.19}$          & $70.53_{\pm0.07}$           & $74.49$          \\
        {SLR (o)}~\cite{mummadi2021test}     & $\textbf{90.32}_{\pm0.27}$ & $76.08_{\pm0.31}$           & $70.61_{\pm0.18}$           & $43.52_{\pm1.15}$           & $70.13$          \\
        Tent~\cite{wang2020tent}             & $ 85.38_{\pm0.43}$         & $ 71.11_{\pm0.58}$          & $ 69.58_{\pm0.21}$          & $70.43_{\pm0.07}$           & $74.12$          \\
        {Tent(o)}~\cite{wang2020tent}        & ${88.70}_{\pm0.34}$        & ${74.86}_{\pm0.46}$         & ${71.14}_{\pm0.27}$         & $46.67_{\pm1.96}$           & $70.34$          \\
        LAME~\cite{boudiaf2022parameter}     & $ 84.77_{\pm0.29}$         & $ 76.66_{\pm0.77}$          & $ 69.46_{\pm0.14}$          & $70.82_{\pm0.12}$           & $75.43$          \\
        \midrule
        DomainAdaptor-T                      & $ {88.74}_{\pm0.30}$       & $ 78.52_{\pm0.57}$          & $ 71.62_{\pm0.14}$          & $ 72.10_{\pm0.09}$          & $77.75$          \\
        DomainAdaptor-SKD                    & $ 88.57_{\pm0.38}$         & $ 79.11_{\pm0.38}$          & $ 71.89_{\pm0.10}$          & $ 72.51_{\pm0.10}$          & $78.02$          \\
        DomainAdaptor-Aug                    & $ 88.45_{\pm0.16}$         & $ \textbf{79.55}_{\pm0.36}$ & $ \textbf{72.84}_{\pm0.05}$ & $ \textbf{73.82}_{\pm0.10}$ & $\textbf{78.67}$ \\
        \bottomrule
      \end{tabular}
      \label{tab_sota}
    }
  \end{center}
  \vspace{-15pt}
\end{table}
% continuous Tent:  0.01 is best 
% PACS : 80.19+-0.69, 80.30+-1.56, 96.66+-0.16, 77.11+-1.68, 83.56+-0.47
% VLCS : 93.23+-1.77, 62.88+-1.14, 70.31+-0.94, 67.08+-1.71, 73.37+-0.31
%  OH  : 58.78+-0.25, 51.21+-0.29, 73.67+-0.15, 74.48+-0.31, 64.54+-0.06
%  MDN : 61.18+-0.82, 31.30+-0.90, 50.72+-4.65, 50.80+-3.35, 48.50+-1.86
% resn50
% PACS : 88.08+-0.67, 84.97+-0.55, 98.17+-0.24, 83.59+-0.33, 88.70+-0.34
% VLCS : 93.96+-0.69, 62.29+-0.62, 74.08+-1.56, 69.10+-1.68, 74.86+-0.46
%  OH  : 67.85+-0.57, 58.03+-0.41, 78.07+-0.49, 80.61+-0.27, 71.14+-0.27
%  MDN : 61.00+-1.62, 30.44+-3.17, 55.16+-3.90, 40.07+-2.98, 46.67+-1.96

% continuous SLR:  0.003 is best 
% PACS : 82.57+-0.38, 83.33+-0.38, 97.04+-0.23, 77.66+-1.07, 85.15+-0.35
% VLCS : 94.73+-1.11, 64.49+-0.57, 69.02+-1.08, 68.29+-1.21, 74.13+-0.58
%  OH  : 59.38+-0.18, 52.37+-0.33, 73.31+-0.33, 73.76+-0.27, 64.71+-0.12
%  MDN : 65.11+-0.45, 27.51+-2.01, 28.44+-3.95, 53.70+-4.81, 43.69+-0.84
% resn50
% PACS : 90.01+-0.68, 87.82+-0.62, 98.40+-0.19, 85.07+-0.35, 90.32+-0.27
% VLCS : 95.14+-0.39, 64.20+-0.90, 73.64+-1.39, 71.33+-1.00, 76.08+-0.31
%  OH  : 68.05+-0.34, 58.01+-0.41, 77.49+-0.17, 78.88+-0.20, 70.61+-0.18
%  MDN : 65.88+-1.30, 35.62+-2.77, 27.17+-1.40, 45.41+-3.30, 43.52+-1.15

Given a trained model, these methods can achieve better performance than the baseline on the PACS dataset, especially for Tent and SLR which employ an online adaption strategy, since a batch of data is enough for the model to produce an approximately accurate statistics estimation on the PACS dataset.  However, when encountered with other datasets (\ie, VLCS, OfficeHome, and MiniDomainNet), they cannot even outperform the baseline. For Tent, SLR, and ARM, the application of AdaBN degrades their performance due to inaccurate estimation of test statistics. Besides, based on AdaBN, ARM and SLR can improve the performance of the PACS dataset, while Tent has little improvement on all datasets. For LAME that does not employ AdaBN, since it considers the relationship between images in a batch, it requires the trained network to make a correct classification for most classes, which could help correctly infer the classes of these images.  However, for VLCS, OfficeHome, and MDN, the trained model has low performance, which is hard for LAME to infer correctly. Besides, OfficeHome and MDN have a large number of classes, which also increases the difficulty of inference.

To better analyze this problem, we calculated the variance of statistics during the training process in \cref{tab_var}. Notably, we observe the highest variance in PACS compared to the other three datasets.
Since training with diverse statistics can effectively enhance the model's robustness to test-time distribution shifts, adopting test statistics that have a different distribution of source statistics can benefit the adaptation process and vice versa. As a result, most methods demonstrate obvious improvement on the PACS dataset but fail to perform well on the other datasets.
Moreover, despite the sensitivity of the trained model to the distribution shift in the training stage, combining the source statistics enables us to obtain a more robust model, as demonstrated by the improved performance of DomainAdaptor. 

By employing AdaMixBN and different GEM variants, our method can mitigate the drawbacks of ARM (requirement of altering training stage), AdaBN (inaccurate statistics estimation), and Tent (small gradient for adaptation) and achieves better performance than the SOTA methods on three datasets.  Specifically, since the predictions on the PACS dataset are not reliable due to the large domain gap, GEM-T can obtain $5.6\%$ performance improvement on the PACS dataset (\ie, $85.04\%$ vs.\ $79.44\%$) by encouraging learning from all classes.
Differently, GEM-SKD is better than GEM-T on other datasets. Because predictions on the dataset with a small domain gap (\ie, VLCS) are more reliable and for the dataset with a large number of classes (\ie, OfficeHome, MDN), only a part of classes are critical for a sample. Thus, learning from all classes is meaningless, and GEM-SKD can perform better.
Since GEM-Aug is based on GEM-SKD, its performance can be further improved based on it by employing more accurate predictions.

\begin{table}[t]
  % \hspace{-20pt}
  \caption{The variance of statistics for different datasets.}
  \vspace{-15pt}
  \begin{center}
    \resizebox{0.6\columnwidth}{!}{
      \begin{tabular}{ l  | c  c c c  }
        \toprule
        \textbf{DataSets} & \textbf{PACS} & \textbf{VLCS} & \textbf{OH} & \textbf{MDN} \\
        \midrule
        Variance          & {1.08}        & 0.05          & 0.78        & 0.43         \\
        \bottomrule
      \end{tabular}
    }
  \end{center}
  \vspace{-5pt}
  \label{tab_var}
\end{table}

%\re{Thank you for the suggestion. We calculated the variance of statistics during the training process, as shown in the table on the right. Notably, we observed the highest variance in PACS. Training with diverse statistics effectively enhances the model's robustness to test-time distribution shifts and vice versa.Furthermore, other methods implemented with AdaBN primarily rely on the statistics of the test batch. Since the model's parameters on PACS are more robust than on other datasets based on previous analysis, it can better adjust the statistics of the test batch. As a result, other methods demonstrate obvious improvement on the PACS dataset but fail to perform well on the other datasets. Moreover, despite the sensitivity of the trained model to the distribution shift in the training stage, combining the source statistics enables us to obtain a robust model, as demonstrated by our proposed AdaMixBN module.}

\begin{table}[t]
  \begin{center}
    \caption{Ablation study of our method with AdaMixBN and GEM-Aug. AdaBN is also included for clearer analysis. T is the shorthand of statistics transformation.}
    \vspace{-5pt}
    \resizebox{\columnwidth}{!}{
      \begin{tabular}{ c | c c c | c c c c }
        \toprule
        \textbf{AdaBN} & \textbf{AdaMixBN w/o T} & \textbf{GEM-Aug} & \textbf{w/ T} & \textbf{PACS} & \textbf{VLCS} & \textbf{OH} & \textbf{MDN} \\
        \midrule
                       &                         &                  &               & $79.44$       & $75.77$       & $64.61$     & $65.12$      \\
        $\checkmark$   &                         &                  &               & $80.44$       & $69.44$       & $63.38$     & $64.20$      \\
                       & $\checkmark$            &                  &               & $83.43$       & $76.62$       & $65.08$     & $65.98$      \\
        %  & $\checkmark$ & only-Aug          &              & 83.98          & 76.90          & 66.16          & 68.10          \\
        % % GEM-T
        % \midrule 
        % $\checkmark$ &              & $\checkmark$ &              & 83.44          & 71.73          & 64.04          & 64.81          \\
        %              & $\checkmark$ & $\checkmark$ &              & 22.23          & 49.30          & 5.96           & 3.29           \\
        %              & $\checkmark$ & $\checkmark$ & $\checkmark$ & \textbf{84.93} & \textbf{78.50} & \textbf{66.73} & \textbf{68.23} \\
        % % GEM-SKD
        % \midrule
        % $\checkmark$ &              & $\checkmark$ &              & 82.18          & 71.76          & 64.09          & 64.99          \\
        %              & $\checkmark$ & $\checkmark$ &              & 59.65          & 44.78          & 44.78          & 25.07               \\
        %              & $\checkmark$ & $\checkmark$ & $\checkmark$ & \textbf{84.37} & \textbf{78.10} & \textbf{65.61} & \textbf{66.42} \\  
        % % GEM-TTA
        % \midrule 
        $\checkmark$   &                         & $\checkmark$     &               & $82.85$       & $72.28$       & $65.37$     & $66.97$      \\
                       & $\checkmark$            & $\checkmark$     &               & $59.47$       & $62.63$       & $44.84$     & $25.68$      \\
                       & $\checkmark$            & $\checkmark$     & $\checkmark$  & $84.93$       & $78.50$       & $66.73$     & $68.23$      \\

        \bottomrule
      \end{tabular}
    }
    \label{tab_ablation}
  \end{center}
  \vspace{-10pt}
\end{table}

\subsection{Ablation Study}

We conducted an ablation study on our DomainAdaptor to investigate the efficacy of each component. The loss is vanilla Entropy Minimization (EM) loss if not mentioned.
%AdaBN is also included for a clearer analysis of our method.

\textbf{AdaBN.} From \cref{tab_ablation}, we can observe that applying AdaBN to the baseline can improve the performance on PACS (\ie, $80.44\%$ vs.\ $79.44\%$) but degrades on other datasets (\eg, $69.44\%$ vs.\ $75.77\%$ on VLCS). We hypothesize that in PACS, the images in the same domain have similar styles, providing more accurate statistics estimation for the batch. Besides, the domain gap in PACS is large, which enhances the effectiveness of AdaBN. On the contrary, in other datasets, the intra-domain images have diverse styles, resulting in inaccurate statistics estimation. Their domain gaps are also relatively small compared to PACS and thus achieve performance lower than the baselines.
% Experiments 

% which normalizes features only with batch statistics,
\textbf{AdaMixBN.} Different from AdaBN, AdaMixBN employs source statistics to ease the inaccurate estimation of statistics. As seen in \cref{tab_ablation}, AdaMixBN can improve the performance largely on PACS data (\ie, $83.43\%$ vs.\ $79.44\%$), and it also can improve on other three datasets. The improvement on PACS shows that by incorporating more source statistics, not only the inaccurate statistics in VLCS and OfficeHome can be accurate, but the already accurate statistics also can be further improved.
%Since using source statistics can largely alleviate the inaccurate statistics estimation problem and improve the baseline.

\textbf{GEM-Aug.} Simply applying EM loss cannot effectively learn from highly confident samples. For instance, Tent obtains $80.59\%$ on PACS in \cref{tab_sota} while AdaBN itself can achieve $80.44\%$ in \cref{tab_ablation}. There is only $0.15\%$ improvement brought by EM loss. Differently, by encouraging further learning from highly confident samples, GEM-Aug loss based on AdaBN can improve more on all four datasets (\eg, $2.41\%$ improvement ($82.85\%$ vs.\ $80.44\%$) on PACS).

\textbf{Transformation.} However, when AdaMixBN is finetuned without the statistics transformation, GEM-Aug loss degrades the performance significantly (\eg, $83.43\%$ vs.\ $59.47\%$ on PACS in \cref{tab_ablation}) due to the weight mismatch problem. Note that not only GEM loss but also other losses that change weight drastically have the same issue, which will be shown in \cref{sec_further}. By transforming the source statistics into affine parameters in AdaMixBN, we can mitigate this issue and the performance can be further improved based on AdaMixBN (\eg, $84.93\%$ vs.\ $83.43\%$ on PACS), which validates the efficacy of our method.

\subsection{Further Analysis}

\label{sec_further}

\begin{table}[t]
  \begin{center}
    \caption{Performance (\%) changes with respect to $\alpha$ for AdaMixBN on different datasets. The best performance of $\alpha$ is marked as \textbf{bold}.}
    \vspace{-5pt}
    \resizebox{\columnwidth}{!}{
      \begin{tabular}{ c|cccccc|c}
        \toprule
        $\boldsymbol{\alpha}$ & $\textbf{0.5}$ & $\textbf{0.6}$ & $\textbf{0.7}$   & $\textbf{0.8}$   & $\textbf{0.9}$   & $\textbf{0.99}$ & \textbf{AdaMixBN} \\
        \midrule
        PACS                  & $83.01$        & $83.41$        & $\textbf{83.64}$ & $83.42$          & $82.35$          & $79.87$         & $83.43$           \\
        VLCS                  & $73.02$        & $74.03$        & $75.04$          & $75.86$          & $\textbf{76.33}$ & $75.87$         & $76.62$           \\
        OH                    & $64.68$        & $64.95$        & $65.20$          & $\textbf{65.35}$ & $65.24$          & $64.72$         & $65.08$           \\
        MDN                   & $65.51$        & $65.77$        & $66.03$          & $\textbf{66.21}$ & $66.11$          & $65.29$         & $65.98$           \\
        \bottomrule
      \end{tabular}
    }
    \label{tab_ratio_alpha}
  \end{center}
  \vspace{-10pt}
\end{table}

\begin{figure*}
  \resizebox{1\linewidth}{!}{

    \begin{subfigure}{0.34\linewidth}
      \begin{center}
        \includegraphics[width=1\linewidth]{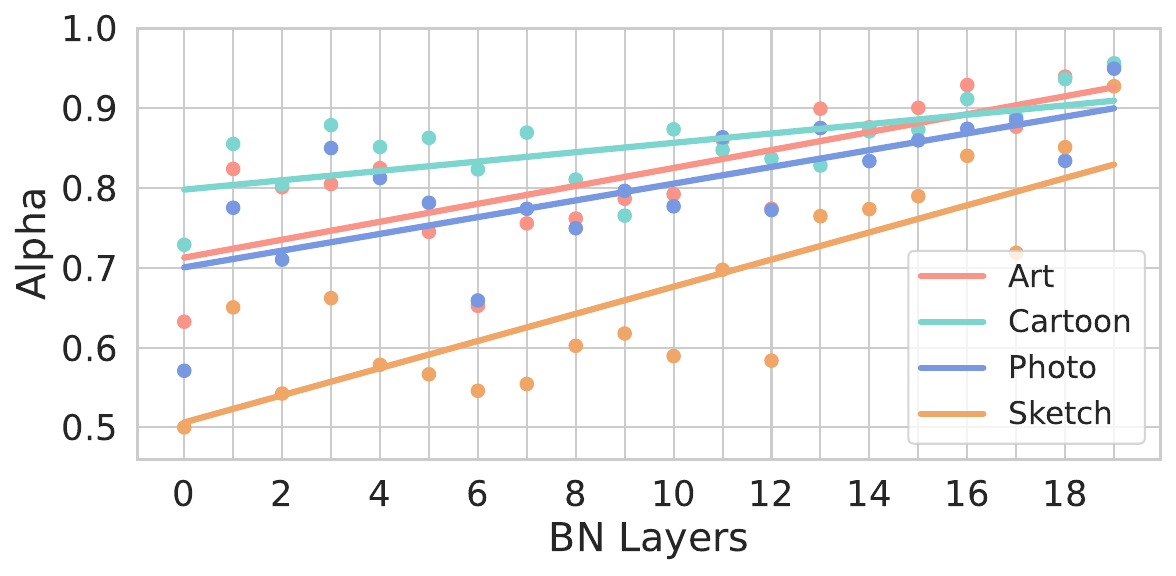}
        \vspace{-15pt}
        \caption{The values of alpha on the PACS dataset.}
        \label{fig_alpha}
      \end{center}
    \end{subfigure}

    \hspace{3pt}

    \begin{subfigure}{0.3\linewidth}
      \vspace{-8pt}
      \center
      \includegraphics[width=1 \linewidth]{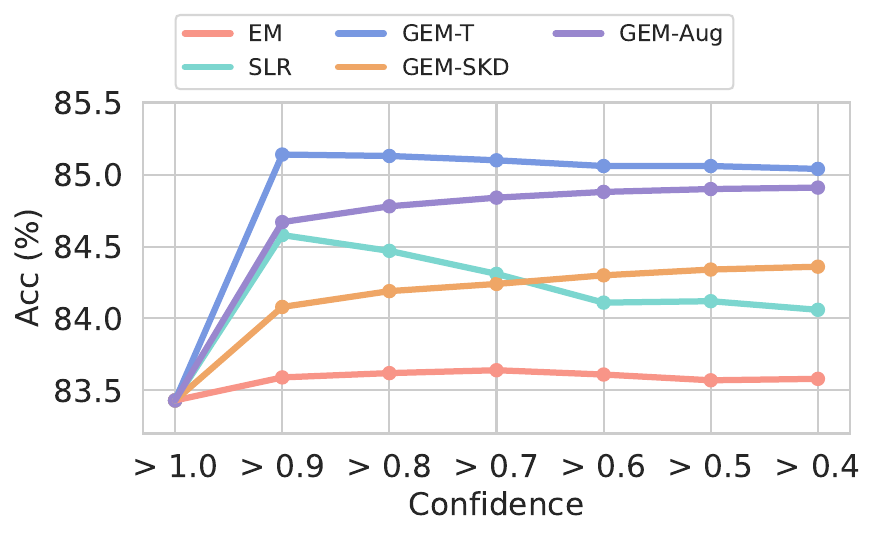}
      \vspace{-15pt}
      \caption{Influence of confident samples.}
      \label{fig_confidence}
    \end{subfigure}

    \hspace{3pt}

    \begin{subfigure}{0.31\linewidth}
      \includegraphics[width=1.0\linewidth]{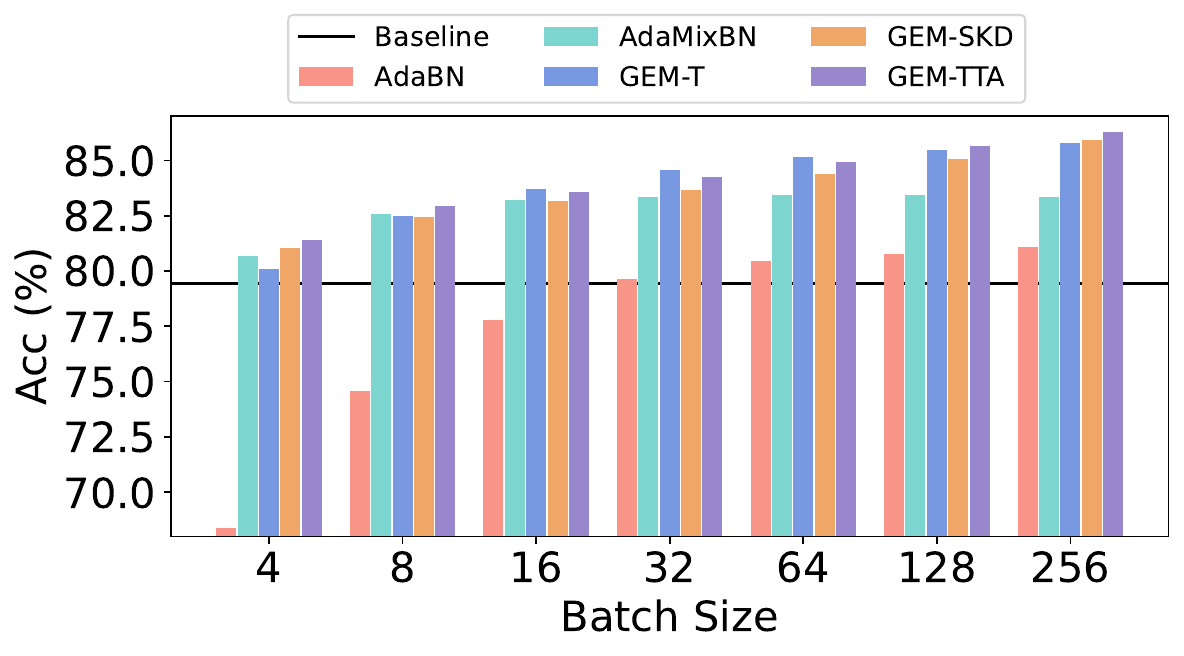}
      \vspace{-15pt}
      \caption{Comparison with different batch sizes.}
      \label{fig_batch_size}
    \end{subfigure}
  }
  \vspace{-10pt}
  \caption{The values of calculated  $\alpha$  in the AdaMixBN (a). The influence of sample (b) and batch size (c) on the performance. } % with their corresponding regression lines, Best viewed in color.
  \vspace{-5pt}
\end{figure*}

\textbf{Whether AdaMixBN learns the optimal $\alpha$.} We propose AdaMixBN to dynamically adjust the coefficient $\alpha$ between source and batch statistics. To validate whether it learns the optimal $\alpha$, we finetune $\alpha$ from $0.5$ to $0.99$ for different datasets to find its optimal value and the corresponding performance. We compare them with the performance produced by the dynamically learned $\alpha$. As shown in \cref{tab_ratio_alpha}, the optimal $\alpha$ in different datasets are different, and AdaMixBN can approximate the performance of these optimal $\alpha$ values by employing a relative distance measure.

% \begin{figure}[t]
%   \begin{center}
%     \includegraphics[width=0.8\linewidth]{imgs/confidence_all.pdf}
%     \vspace{-0.4cm}
%     \caption{The performance of finetuning the trained model with different confident samples.}
%     \label{fig_confidence}
%   \end{center}
%   \vspace{-15pt}
% \end{figure}

\label{sec_degradation}
\textbf{Performance degradation with different $\alpha$.} To investigate whether the performance degradation phenomenon is only caused by GEM loss or also other losses and how $\alpha$ in AdaMixBN influences the degree of degradation, we experiment with three losses (\ie, EM loss, SLR loss, and GEM-Aug loss) equipped with AdaMixBN on PACS with different $\alpha$. As shown from \cref{tab_degradation}, when $\alpha$ becomes large, not only does using GEM-Aug loss cause the performance drop but also SLR degrades the performance.
Differently, the performance of EM loss does not drop significantly because both GEM loss and SLR loss change model weights drastically, while EM loss makes little changes to the model weights due to the small gradients it produces, resulting in a similar performance to AdaMixBN (\eg,  when $\alpha=0.5$, $83.01\%$ vs. $83.14\%$).
Besides, we find that with more source statistics mixed, the performance drops more significantly. This is caused by the mismatch between fixed source statistics and finetuned model weights and can be addressed with the statistics transformation. As shown in the last line in \cref{tab_degradation}, after we applying the statistics transformation operation, GEM-Aug loss improves the performance consistently instead of hurting the performance.

\begin{table}[t]
  \begin{center}
    \caption{Performance (\%) degradation with respect to $\alpha$ using different losses in AdaMixBN.}
    \resizebox{0.9\columnwidth}{!}{
      \begin{tabular}{ c|cccccc}
        \toprule
        $\boldsymbol{\alpha}$ & $\textbf{0.1}$ & $\textbf{0.3}$ & $\textbf{0.5}$ & $\textbf{0.7}$ & $\textbf{0.9}$ & $\textbf{0.99}$ \\
        \midrule
        AdaMixBN              & $80.94$        & $82.01$        & $83.01$        & $83.64$        & $82.35$        & $79.87$         \\
        \midrule
        \multicolumn{7}{c}{Finetune \textit{without} statistics transformation}                                                      \\
        \midrule
        EM                    & $81.05$        & $82.22$        & $83.14$        & $83.79$        & $82.41$        & $79.10$         \\
        SLR                   & $81.98$        & $83.10$        & $84.17$        & $83.53$        & $65.66$        & $43.08$         \\
        % GEM-T    & 83.86 & 84.13 & 82.76 & 74.04 & 32.13 & 18.66 \\ 
        % GEM-SKD & 82.32 & 81.47 & 74.20 & 48.92 & 22.47 & 19.47 \\
        GEM-Aug               & $83.07$        & $82.06$        & $74.40$        & $48.43$        & $22.80$        & $19.77$         \\
        \midrule
        \multicolumn{7}{c}{Finetune \textit{with} statistics transformation}                                                         \\
        \midrule
        % GEM-T         & 83.75 & 84.45 & 85.17 & 85.54 & 84.38 & 82.30 \\
        % GEM-SKD       & 82.61 & 83.38 & 84.08 & 84.44 & 83.93 & 82.11 \\
        GEM-Aug               & $83.32$        & $84.07$        & $84.61$        & $85.08$        & $82.49$        & $82.49$         \\
        \bottomrule
      \end{tabular}
    }
    \label{tab_degradation}
  \end{center}
  \vspace{-15pt}   
\end{table}

\textbf{The values of dynamic $\alpha$.} In \cref{sec_alpha} we argue that lower layers contain more domain-specific information while higher layers contain less. Therefore a dynamic mixture coefficient is proposed for statistic fusing. To investigate whether this is the case, we plot $\alpha$ obtained by \cref{eq_alpha} in different layers of BN  and their regression lines on the PACS dataset. As seen from \cref{fig_alpha}, in all four domains, $\alpha$ increases as the layer becomes higher, which indicates that source statistics in higher layers are more transferable and should be incorporated with a larger value of $\alpha$.

\textbf{Contributions of different confident samples.} To investigate whether our proposed GEM loss can better learn from highly confident samples, we compare different losses by finetuning AdaMixBN with samples whose confidence predicted by the model is higher than a threshold. As shown in \cref{fig_confidence}, with more low-confident samples incorporated, the performance of EM loss only slightly improves and then decreases due to the noise introduced by the more low-confident samples. While SLR loss can better exploit samples with confidence larger than $0.9$, it decreases a lot when more low-confident samples are added since it cannot inhibit the noise introduced by the low-confident samples. Different from these two losses, by only finetuning samples with confidence higher than $0.9$, our three losses can achieve higher performance than EM loss. Besides, by employing the temperature to soften the probability, noise can be reduced and our two losses (\ie, GEM-Aug and GEM-SKD) can all benefit from more low-confident samples and achieve better performance. The performance of GEM-T drops slightly since it encourages the model to make different predictions, which is more sensitive to batch noise.

\textbf{The influence of batch size.} Since both AdaMixBN and the adaptation process are closely related to the batch size, we compare AdaBN, AdaMixBN, and our proposed three losses to see how the batch size influences their performances. As shown in \cref{fig_batch_size}, with a batch size of $4$, the performance of AdaBN drops drastically from $79.44\%$ to $68.40\%$ because the statistics estimated from only $4$ images are very inaccurate and cannot effectively normalize the features. On the contrary, our proposed AdaMixBN can still achieve higher performance than the baseline by incorporating more source statistics. As the number of images increases, the performance can be further improved and then plateaus, which means the statistics are accurate and can not be improved. Although the performance of AdaBN also increases, AdaMixBN can still outperform it with the source statistics. When more unlabeled images are utilized, the performances of adaptation with different losses also increase. We hypothesize that sample diversity plays a significant role in the adaptation process. With more samples incorporated, the noise produced by incorrectly predicted samples is alleviated, and the optimization direction toward the global minimum can be more accurate.

\label{exp:continous}

\begin{table}[t]
  \begin{center}
    \caption{{Performance (\%) comparison with few data.}}
    \vspace{-5pt}
    \resizebox{\columnwidth}{!}{
      \begin{tabular}{ c|ccccccc}
        \toprule
        \textbf{Subset Size} & $\textbf{64}$ & $\textbf{128}$ & $\textbf{256}$ & $\textbf{512}$ & $\textbf{1024}$ & $\textbf{2048}$ & $\textbf{4096}$ \\
        \midrule
        Tent                 & 80.79         & 81.00          & 81.24          & 81.75          & 82.41           & 82.95           & 83.52           \\
        SLR                  & 81.33         & 81.99          & 82.55          & 83.46          & 84.26           & 84.79           & 85.15           \\
        DomainAdaptor-T      & 85.04         & 85.04          & 85.04          & 85.04          & 85.04           & 85.04           & 85.04           \\
        \bottomrule
      \end{tabular}
    }
    \label{tab_small}
  \end{center}
  \vspace{-10pt}
\end{table}

% \begin{figure}[t]
%   \centering 
%   \includegraphics[width=0.8\linewidth]{imgs/small_dataset.pdf}
%     \vspace{-0.4cm}
%   \caption{The performance comparison when the model is trained with a small dataset.}
%   \label{fig_small}
% \end{figure}

{\textbf{Comparison with few test data.} To compare with other methods under limited test data, we divide the original dataset into several subsets of the same size and evaluate all the methods on each subset, with model parameters reset before finetuning on a new subset. The adaption batch size is $64$. The resulting accuracies are then aggregated across all subsets. As shown from \cref{tab_small}, the performances of both Tent and SLR are low when the subset size is small, indicating their poor adaptation ability in the few-data scenario. Differently, DomainAdaptor can achieve consistent improvement regardless of subset size due to its ability to perform adaptation with only a single batch.}

\begin{table}[t]
  \begin{center}
    \caption{Performance (\%) of applying DomainAdaptor to previously pretrained models.}
    \vspace{-5pt}
    \resizebox{\columnwidth}{!}{
      \begin{tabular}{ l|cccc|c}
        \toprule
                                     & \textbf{P} & \textbf{A} & \textbf{C} & \textbf{S} & \textbf{Avg.} \\
        \midrule
        DeepAll                      & $96.44$    & $78.25$    & $75.57$    & $67.48$    & $79.44$       \\
        + DomainAdaptor-T            & $97.40$    & $82.51$    & $80.65$    & $79.59$    & $ 85.04$      \\
        + DomainAdaptor-SKD          & $97.01$    & $82.22$    & $80.46$    & $77.78$    & $ 84.37$      \\
        + DomainAdaptor-Aug          & $97.20$    & $83.04$    & $80.20$    & $79.28$    & $ 84.93$      \\
        \midrule
        MLDG~\cite{li2018learning}   & $94.91$    & $80.59$    & $76.20$    & $77.65$    & $82.34$       \\
        + DomainAdaptor-T            & $96.35$    & $82.84$    & $81.28$    & $80.96$    & $85.36$       \\
        + DomainAdaptor-SKD          & $96.36$    & $83.04$    & $81.41$    & $81.02$    & $85.46$       \\
        + DomainAdaptor-Aug          & $96.16$    & $82.95$    & $81.37$    & $81.54$    & $85.50$       \\
        \midrule
        FSDCL~\cite{wang2022feature} & $95.63$    & $85.30$    & $81.31$    & $81.17$    & $85.85$       \\
        + DomainAdaptor-T            & $96.65$    & $87.01$    & $83.53$    & $81.04$    & $87.06$       \\
        + DomainAdaptor-SKD          & $96.65$    & $87.40$    & $83.66$    & $79.28$    & $86.75$       \\
        + DomainAdaptor-Aug          & $96.53$    & $86.57$    & $83.70$    & $80.89$    & $86.92$       \\
        \midrule
        MVRML~\cite{zhang2022mvdg}   & $95.29$    & $85.20$    & $79.97$    & $83.11$    & $85.89$       \\
        + DomainAdaptor-T            & $96.18$    & $85.99$    & $83.98$    & $83.73$    & $87.47$       \\
        + DomainAdaptor-SKD          & $96.23$    & $85.92$    & $83.86$    & $83.59$    & $87.40$       \\
        + DomainAdaptor-Aug          & $96.41$    & $86.69$    & $84.33$    & $85.05$    & $88.12$       \\
        \bottomrule
      \end{tabular}
    }
    \label{tab_trained}
  \end{center}
  \vspace{-15pt}
\end{table}

\textbf{The performance of being applied to the trained model.} Since our method does not require altering the training stage and thus can be applied to any trained CNN models, we apply it to several trained recent DG SOTA models on PACS. As seen from \cref{tab_trained}, our method can significantly improve the baseline method's performance (\eg, $85.04\%$ vs. $79.44\%$). When applied to existing DG SOTA models, our method can make further improvements (\eg, $88.12\%$ vs. $85.89\%$ for MVRML) without the requirement of altering the training stage and retraining the model.

\begin{table}[t]
  \begin{center}
    \caption{Performance (\%) comparison of adapting with different self-supervised losses or different statistics. The baselines of rotation and jigsaw losses are denoted as Base.}
    \vspace{-5pt}
    \renewcommand\arraystretch{1.1}
    \resizebox{1.0\columnwidth}{!}{
      \begin{tabular}{ l  | c  c c c | c }
        \toprule
                              & \textbf{PACS}       & \textbf{VLCS}       & \textbf{OH}         & \textbf{MDN}        & \textbf{Avg.} \\
        \midrule
        DeepAll               & $79.44_{\pm0.44}$   & $75.77_{\pm0.29}$   & $64.61_{\pm0.18}$   & $65.12_{\pm0.11}$   & $71.24$       \\
        \midrule
        Rot (Base)            & $80.91_{\pm0.67}$   & $75.02_{\pm0.38}$   & $64.24_{\pm0.15}$   & $64.65_{\pm0.09}$   & $71.22$       \\
        + RotLoss           & $81.26_{\pm0.46}$   & $69.82_{\pm0.61}$   & $63.00_{\pm0.24}$   & $63.50_{\pm0.09}$   & $69.39$       \\
        + GEM-Aug             & $83.16_{\pm0.76}$   & $72.77_{\pm0.63}$   & $63.87_{\pm0.32}$   & $65.41_{\pm0.10}$   & $71.29$       \\
        \midrule
        Jig (Base)            & $78.60_{\pm0.73}$   & $74.91_{\pm0.60}$   & $61.70_{\pm0.58}$   & $62.55_{\pm0.10}$   & $69.43$       \\
        + JigLoss             & $80.24_{\pm0.21}$   & $70.53_{\pm1.07}$   & $62.47_{\pm0.44}$   & $63.85_{\pm0.11}$   & $69.25$       \\
        + GEM-Aug             & $84.51_{\pm0.40}$   & $73.64_{\pm1.19}$   & $63.95_{\pm0.33}$   & $64.92_{\pm0.10}$   & $71.75$       \\
        \midrule
        AdaBN                 & $80.44_{\pm0.29}$   & $69.44_{\pm0.48}$   & $63.38_{\pm0.12}$   & $64.20_{\pm0.11}$   & $69.37$       \\
        AdaBN (src+test)      & $81.80_{\pm0.45}$   & $76.11_{\pm0.31}$   & $65.26_{\pm0.20}$   & $65.89_{\pm0.12}$   & $72.27$       \\
        \midrule
        AdaMixBN ($\alpha$)   & $83.43_{\pm0.23}$   & $76.62_{\pm0.27}$   & $65.08_{\pm0.12}$   & $65.98_{\pm0.10}$   & $72.78$       \\
        AdaMixBN ($1-\alpha$) & $81.31_{\pm0.27}$   & $69.51_{\pm0.50}$   & $63.47_{\pm0.13}$   & $64.36_{\pm0.10}$   & $69.66$       \\
        \midrule
        DomainAdaptor-Aug     & ${84.93}_{\pm0.19}$ & ${78.50}_{\pm0.22}$ & ${66.73}_{\pm0.25}$ & ${68.23}_{\pm0.08}$ & ${74.60}$     \\
        \bottomrule
      \end{tabular}
      \label{tab_compare}
    }
  \end{center}
  \vspace{-15pt}
\end{table}

\textbf{Comparison to self-supervised losses.} To determine the effectiveness of our method compared to previous self-supervised losses, we conducted a comparison with rotation~\cite{sun2020test} and jigsaw~\cite{carlucci2019domain} loss. Since these self-supervised losses require an extra classification head, to assess their performance, we re-train the DeepAll with a new head (denoted as Base) and report the results with a single finetuning for each batch.  As shown in \cref{tab_compare}, these losses also do not yield significant performance improvement since they cannot produce a sufficiently large gradient for weight updating. Differently, benefiting from GEM loss, our method still can produce a substantial performance boost.

\textbf{Normalization with different statistics.}  Since calculating the statistics with source and target data may serve as an upper bound, we recalculate the statistics, as shown in \cref{tab_compare} (denoted as AdaBN (src+test)). While we observe an improvement compared to AdaBN, it only results in marginal improvement over the baseline. Differently, our proposed AdaMixBN consistently achieves better performance than the baseline. We hypothesize that AdaMixBN benefits from the dynamically mixing statistics between the source and target statistics from different layers, which cannot be achieved by simply accumulating the statistics. Besides, we also adopt $1-\alpha$ during mixing to verify whether the similarity-based distance works. As shown in \cref{tab_compare}, it degrades significantly compared to using $\alpha$, which validates the effectiveness of our proposed measurement.

\section{Conclusion}
In this paper, we proposed DomainAdaptor to deal with the fully test-time adaptation problem for any incoming test batch with a large domain gap. It consists of AdaMixBN and Generalized Entropy Minimization loss to effectively exploit unlabeled test data. Specifically, AdaMixBN dynamically fuses statistics between source and test batch statistics for an accurate statistics estimation and Generalized Entropy Minimization loss effectively enhances the adaptation ability of the AdaMixBN module. Besides, a statistics transformation operation was incorporated to prevent performance degradation in AdaMixBN. Extensive experiments validated the effectiveness of our method.

$\\$
\noindent \textbf{Acknowledgment}: This work is supported by NSFC Program (62222604, 62206052, 62192783), China Postdoctoral Science Foundation Project (2023T160100), Jiangsu Natural Science Foundation Project (BK20210224), and CCF-Lenovo Bule Ocean Research Fund.

{\small
\bibliographystyle{ieee_fullname}
\bibliography{egbib}
}
\clearpage

\appendix

\section{Supplementary Material}

In the Supplementary Material, we first provide the detailed derivation of the gradient of generalized entropy minimization loss. Following it, the implementation details are provided. Then we apply our method to a strong baseline ERM and a SOTA method SWAD on the  DomainBed benchmark to further validate its wide application. Moreover, we show that previous methods heavily rely on continuous adaptation, which may perform poorly in multi-domain scenarios. Finally, we provide the full results of the performance comparison in Sec. 4.1 in the main text.

\section{Formula Derivations}

\textbf{The derivation of statistics transformation.} Given feature map $x$, the source statistics $(\mu_s, \sigma_s)$ and target statistics $(\mu_t, \sigma_t)$ for each normalization layer, where the layer indices are omitted for simplicity. To eliminate the negative effect of source statistics, we propose to transform the source statistics into the affine parameter in the normalization layer with the following formula:
\begin{align}
    & \frac{x-(\alpha\mu_s+(1-\alpha)\mu_t)}{\sqrt{\alpha\sigma_s^2+(1-\alpha)\sigma_t^2}}\cdot\gamma+\beta                          \\
  = & \frac{x-(\alpha\mu_s+(1-\alpha)\mu_t)}{\sigma_t}\cdot\frac{\sigma_t}{\sqrt{\alpha\sigma_s^2+(1-\alpha)\sigma_t^2}}\gamma+\beta \\
  = & \frac{x-(\alpha\mu_s+(1-\alpha)\mu_t)}{\sigma_t}\cdot\gamma'+\beta                                                             \\
  = & (\frac{x-\mu_t}{\sigma_t}+\frac{\alpha(\mu_t-\mu_s)}{\sigma_t})\cdot\gamma'+\beta                                              \\
  = & \frac{x-\mu_t}{\sigma_t}\cdot\gamma' + \frac{\alpha(\mu_t-\mu_s)}{\sigma_t}\cdot\gamma'+\beta                                  \\
  = & \frac{x-\mu_t}{\sigma_t}\cdot\gamma' + \beta',
\end{align} where we set
\begin{align}
  \gamma' & = \frac{\sigma_t}{\sqrt{\alpha \sigma_s^2+(1-\alpha)\sigma^2_t}}\gamma + \beta, \\
  \beta'  & =  \frac{\alpha(\mu_t-\mu_s)}{\sigma_t} \gamma' + \beta.
\end{align}
This process can be seen as a re-initialization of the affine parameter, which is done before the finetuning process, that is, we finetune the transformed parameters instead of the original parameters.

\begin{table*}[t]
  \begin{center}
    \vspace{-0.3cm}
    % \caption{The average running time of a batch (64 images) on Art.} 
    \caption{The running time (ms) on the Art domain with 2048 images.}
    \vspace{-0.2cm}
    \resizebox{0.8\linewidth}{!}{
      \begin{tabular}{ c|cccccc|cc}
        \toprule
                  & \textbf{Non-Adapted} & \textbf{AdaBN} & \textbf{LAME}  & \textbf{ARM}    & \textbf{SLR}    & \textbf{Tent}   & \textbf{AdaMixBN} & \textbf{DomainAdaptor-T} \\
        \midrule
        % Time (s) & 1.10        & 1.19  & 1.34  & 5.63  & 4.98       & 4.87  & 1.34     & 6.23            \\
        Time (ms) & 34.4        & 37.19 & 41.88 & 175.94 & 155.63 & 152.19 & 41.88    & 194.69          \\
        Acc (\%)  & 78.25       & 76.36 & 80.05 & 81.02  & 81.66  & 81.06  & 80.81    & 82.51  \\
        \bottomrule
      \end{tabular}
    }
    \label{tab_time}
  \end{center}
  \vspace{-0.7cm}
\end{table*}
\textbf{The derivation of Generalized Entropy Minimization (GEM) loss.} The GEM loss is
\begin{align}
  L     & =-\sum p_{i} \log q_{i},                                                                              \\
  p_{i} & = \frac{e^{z_i/\tau_1}}{\sum_j e^{e_j/\tau_1}}, q_{i} = \frac{e^{z_i/\tau_2}}{\sum_j e^{e_j/\tau_2}}.
\end{align}
And the gradient of $p_i$ with respect to the logits $z_i$ is:
\begin{align}
  \frac{\partial p_i}{\partial z_i} = \Bigl\{
  \begin{array}{lr}
    p_i(1-p_i), & \text{for } 0\leq n\leq 1   \\
    -p_ip_j,    & \text{for } 0\leq n\leq 1 .
  \end{array}
\end{align}
Then the gradient of logit $z_k$ of class $k$ can be obtained by the following formula:
\begingroup
\allowdisplaybreaks
\begin{align}
  \frac{\partial L}{\partial z_{k}}
   & =-\Bigr[\frac{\partial p_{k}  \log q_{k}}{\partial z_{k}} + \frac{\partial \sum_{j\not=k} p_{j} \log q_{j}}{\partial z_{k}}\Bigr]                                          \\
   & =-\Bigr[\frac{\partial p_{k}}{\partial z_{k}}  \log q_{k} +p_{k}  \frac{1}{q_{k}}  \frac{\partial q_{k}}{\partial z_{k}}    \notag                                         \\
   & \phantom{=-\Bigr[\frac{\partial p_{k}}{\partial }  }
  + \sum_{j \neq k}\left(\frac{\partial p_{j}}{\partial z_{k}}  \log q_{j} + p_{j} \frac{1}{q_{j}}  \frac{\partial q_{j}}{\partial z_{k}}\right)\Bigr]                          \\
   & =-\Bigr[\frac{1}{\tau_1}  p_{k} \left(1-p_{k}\right)  \log q_{k}+\frac{p_{k}}{q_{k}}  \frac{1}{\tau_2} q_{k}\left(1-q_{k}\right)                                           \\
   & \phantom{-\Bigr[\frac{1}{\tau_1 p_{k} } } +\sum_{j \neq k}\left(-\frac{1}{\tau_1}  p_{j} p_{k}  \log q_{j}-\frac{p_{j}}{q_{j}}  \frac{1}{\tau_2} q_{j}  q_{k}\right)\Bigr] \\
   & =-\Bigr[\frac{1}{\tau_1} p_{k} \left(1-p_{k}\right)  \log q_{k}+\frac{1}{\tau_2} p_{k}\left(1-q_{k}\right)              \notag                                             \\
   & \phantom{ =-\Bigr[\frac{1}{\tau_1} p_{k} } - \sum_{j \neq k}\left(\frac{1}{t} p_{j}  p_{k}  \log q_{j}+\frac{1}{\tau_2} p_{j}  q_{k}\right)\Bigr]                          \\
   & =-\Bigr[\frac{1}{\tau_1} p_{k}  \log q_{k}+\frac{1}{\tau_2} p_{k} \notag                                                                                                   \\
   & \phantom{ =-\Bigr[\frac{1}{\tau_1} p_{k} } -\sum_{j}\left(\frac{1}{\tau_1} p_{j}  p_{k}  \log q_{j}+\frac{1}{\tau_2} p_{j}  q_{k}\right)\Bigr]                             \\
   & =-\Bigr[\frac{1}{\tau_1} p_{k}  \log q_{k}+\frac{1}{\tau_2} p_{k} \notag                                                                                                   \\
   & \phantom{ =-\Bigr[\frac{1}{\tau_1} p_{k} } -\frac{1}{\tau_1} p_{k}  \sum_{j} p_{j}  \log q_{j}-\frac{1}{\tau_2} q_{k}\Bigr]                                                \\
   & =-\Bigr[\frac{1}{\tau_2} \left(p_{k}-q_{k}\right) \notag                                                                                                                   \\
   & \phantom{ =-\Bigr[\frac{1}{\tau_1} p_{k} } +\frac{p_{k}}{\tau_1} \left(\log q_{k}-\sum_{j} p_{j}  \log q_{j}\right)\Bigr].
  % & =-\Bigr[\frac{1}{\tau_2} \left(p_{k}-q_{k}\right)+\frac{p_{k}}{\tau_1} \left(\log q_{k}+L\right)\Bigr]
\end{align}
\endgroup 

When the gradient of $p_i$ is detached, the gradient of logit $z_k$ becomes:
\begin{align}
  \frac{\partial L}{\partial z_{k}} & =-\sum_{i} \frac{p_{i}}{q_{i}}  \frac{\partial q_{i}}{\partial z_{k}} \ \ \ \ \ \ \text{(no gradient to $p_i$)} \\
  % &=-\left(\frac{1}{\tau_2}  \frac{p_{k}}{q_{k}}  q_{k} \left(1-q_{k}\right)-\frac{1}{\tau_2}  \sum_{j \neq k} \frac{p_{j}}{q_{j}} q_{k}  q_{j}\right) \\
                                    & =-\frac{1}{\tau_2} \left(p_{k}-p_{k}  q_{k}-\sum_{j \neq k} p_{j}  q_{k}\right)                                 \\
                                    & =-\frac{1}{\tau_2} \left(p_{k}-\sum_{j} p_{j}  q_{k}\right)=-\frac{1}{\tau_2} \left(p_{k}-q_{k}\right),
\end{align}
which takes the same form of knowledge distillation~\cite{hinton2015distilling}.

% \clearpage 

\section{Implementation Details}

In all experiments, we adopt Resnet-18 and Resnet-50~\cite{he2016deep} models trained with ERM (\ie, simply aggregate all source data) as our baseline. During adaptation, the learning rate of the SGD optimizer is set to $1e-3$ without momentum. The default batch size is 64 and test images are resized to $224\times 224$ without other augmentation. For GEM-Aug, we adopt weak augmentations that consist of a random crop with a scale range of $[0.8, 1]$ and a random flip with a probability of $0.5$. The test order of samples is fixed for a fair comparison to each method.

\section{More Experiments}

\subsection{Time cost comparison} 
We have added the run-time comparison of our method and previous methods with a batch size of 64 on the Art domain. The experiments are done on an RTX2080 GPU. As shown in \cref{tab_time}, with a little computational overhead, our method could achieve better performance.

\subsection{Experiments on DomainBed}

We apply our method to ERM and SWAD trained on DomainBed on three datasets (\ie, PACS~\cite{li2017deeper}, VLCS~\cite{torralba2011unbiased} and OfficeHome~\cite{venkateswara2017deep}). The checkpoint of ERM is selected in the last iteration and SWAD is the ensembled version of ERM. We test the performance of these methods on the whole leave-out domain with a batch size of 64 and a learning rate of 0.05. The results are averaged by three independent runs. By applying our method to these two methods, although SWAD could achieve strong performance on DomainBed, we still could improve on it by fully exploiting the information in the test batch for adaptation.

\begin{table}[h]
  \begin{center}
    \renewcommand\arraystretch{1.1}
    \caption{Performance (\%) comparison to ERM and SWAD on the DomainBed benchmark.}  
    \resizebox{\columnwidth}{!}{
      \begin{tabular}{ l|ccc|c}
        \toprule
                           & \textbf{PACS} & \textbf{VLCS} & \textbf{OfficeHome} & \textbf{Avg.} \\
        \midrule
        ERM                & 83.32         & 75.51         & 65.30               & 74.71         \\
        +MixAdaBN          & 85.76         & 76.00         & 66.03               & 75.93         \\
        +DomainAdaptor-T   & 86.61         & 76.60         & 66.70               & 76.63         \\
        +DomainAdaptor-SKD & 86.31         & 76.71         & 66.60               & 76.54         \\ 
        +DomainAdaptor-Aug & 86.68         & 77.09         & 67.51               & 77.09         \\ 
        \midrule
        ERM+SWAD           & 86.77         & 77.62         & 70.31               & 78.23         \\
        +MixAdaBN          & 88.88         & 78.83         & 70.82               & 79.51         \\
        +DomainAdaptor-T   & 89.42         & 79.02         & 70.90               & 79.78         \\
        +DomainAdaptor-SKD & 89.30         & 79.21         & 71.03               & 79.85         \\
        +DomainAdaptor-Aug & 89.62         & 79.58         & 71.71               & 80.30         \\
        \bottomrule
      \end{tabular}
    }
    \label{tab:tent}
  \end{center}
  \vspace{-0.4cm}
\end{table}

\subsection{Comparison to continuous adaptation}
% we argue that ``\textit{in addition to the estimation of the inaccurate statistics, their continuous adaptation could suffer from catastrophic forgetting and continuous performance degradation~\cite{wang2022continual} when multiple test domains exist.}" 
Since our method only requires a single finetuning iteration by fully exploiting the information of a test batch, which differs from the previous test-time adaptation methods~\cite{wang2020tent,mummadi2021test,boudiaf2022parameter} that have a large demand of data by employing online updating. To further demonstrate the poor adaptation ability of previous test-time adaptation method, we conduct several experiments to verify that 1) these methods rely on the continuous adaptation and cannot effectively exploit the current batch data in \cref{tab:tent}; 2) the performance degradation occurs when the original prediction is inaccurate in~\cref{tab:continual} or multiple domains exist in~\cref {tab:continual_multiple}.

\begin{table}[h]
  \begin{center}
    \renewcommand\arraystretch{1.1}
    \caption{The performance (\%) of Tent without online updating weights or without the momentum term in SGD.}
    \resizebox{\columnwidth}{!}{
      \begin{tabular}{ l|cccc|c}
        \toprule
                          & \textbf{P} & \textbf{A} & \textbf{C} & \textbf{S} & \textbf{Avg.}      \\
        \midrule
        baseline          & $96.44$    & $78.25$    & $75.57$    & $67.48$    & $ 79.44_{\pm0.44}$ \\
        \midrule
        Tent w/o both     & $95.70$    & $76.38$    & $78.75$    & $71.07$    & $ 80.47_{\pm0.28}$ \\
        Tent w/o Online   & $95.80$    & $76.69$    & $78.98$    & $71.69$    & $ 80.79_{\pm0.19}$ \\
        Tent w/o Momentum & $95.87$    & $77.05$    & $79.33$    & $73.43$    & $ 81.42_{\pm0.17}$ \\
        Tent              & $96.42$    & $79.39$    & $80.86$    & $77.44$    & $ 83.53_{\pm0.42}$ \\
        \bottomrule
      \end{tabular}
    }
    \label{tab:tent}
  \end{center}
  \vspace{-0.4cm}
\end{table}

\textbf{Tent relies on continuous finetuning.} We argue that the success of Tent relies on two critical factors: the continuous finetuning and the momentum term in the optimizer. To verify it, we conduct an ablation study by only updating the model weights online without momentum in the optimizer or only enabling the momentum without online updating the model weights. As shown in \cref{tab:tent}, both momentum term and online updating could improve the performance of Tent (\ie, 83.53\% vs.\ 80.79\%, and 83.53\% vs.\ 81.42\%). The online updating could preserve the learned knowledge from previous batches, while the momentum term could utilize the history gradients to guide current gradients. Also, we could find that online updating is more important since all of the learned knowledge is kept in the finetuned weights. However, when both are missing, there is a little improvement of Tent, which is actually owing to AdaBN~\cite{li2016revisiting} (80.44\% on PACS) equipped in Tent, as mentioned before. Therefore, Tent that adapts only once cannot effectively exploit the unlabeled batch.

\begin{table}[t]
  \begin{center}
    \renewcommand\arraystretch{1.1}
    \caption{The continuous adaptation with Tent and its variants that continuously update the source statistics with momentum $m$. Best performance (\%) is \textbf{bolded} for Tent.}
    \resizebox{0.75\columnwidth}{!}{
      \begin{tabular}{ l|ccccccccccc}
        \toprule
        \textbf{Method}   & \textbf{PACS}    & \textbf{VLCS}    & \textbf{OH}      & \textbf{MDN}     \\ 
        \midrule
        \multicolumn{5}{c}{\textit{Adaptation only once}}                                        \\
        \midrule
        DeepAll           & $79.44$          & $75.77 $         & $64.61$          & $65.12$          \\
        Tent              & $80.59$          & $69.69$          & $63.58$          & $64.37$          \\
        DomainAdaptor-T   & $85.04$          & $77.54$          & $65.39$          & $66.39$          \\
        DomainAdaptor-SKD & $84.37$          & $78.10$          & $65.61$          & $66.42$          \\
        DomainAdaptor-Aug & $84.93$          & $78.50$          & $66.73$          & $68.23$          \\
        \midrule
        \multicolumn{5}{c}{\textit{Continuous adaptation}}                                         \\
        \midrule
        Tent, $m=0.0$     & $49.42$          & $58.69$          & $16.08$          & $1.36 $          \\
        Tent, $m=0.1$     & $80.47$          & $65.34$          & $51.09$          & $4.71 $          \\
        Tent, $m=0.5$     & $\textbf{83.89}$ & $73.07$          & $\textbf{64.55}$ & $26.54$          \\
        Tent, $m=0.9$     & $83.60$          & $\textbf{73.42}$ & $\textbf{64.55}$ & $47.44$          \\
        Tent, $m=1.0$     & $83.53$          & $73.37$          & $64.52$          & $\textbf{48.23}$ \\
        \bottomrule
      \end{tabular}
    }
    \label{tab:continual}
  \end{center}
  \vspace{-0.4cm}
\end{table}

\begin{table}[t]
  \begin{center}
    \renewcommand\arraystretch{1.1}
    \caption{The performance (\%) of continuous adaptation on PACS datasets. The baseline is trained on the Photo domain and the best performance is \textbf{bolded}.
    % (c) means the consecutive batches are sampled from the same domains, while (s) means these batches are randomly sampled from different domains.
    }
    \resizebox{0.9\columnwidth}{!}{
      \begin{tabular}{l | c c c | c c c | c c }
        % shuffled=36.33, not shuffled=31.33, the data from the same domain would continuously degrade the
        \toprule
                 & \textbf{A}     & \textbf{C}     & \textbf{S}     & \textbf{AC}    & \textbf{AS}    & \textbf{CS}    & \textbf{ACS}   \\
        \midrule
        Baseline & $59.08$        & $25.78$        & $29.84$        & $41.45$        & $39.90$        & $28.33$        & $35.96$        \\
        % 62.79 & 47.40 & 47.46 & 54.64 & 52.74 & 47.44 & 51.25 & 51.25 \\
        Tent     & \textbf{67.53} & \textbf{62.20} & $43.70$        & $48.18$        & $36.79$        & $31.28$        & $36.33$        \\
        Ours     & $64.45$        & $49.70$        & \textbf{50.79} & \textbf{56.64} & \textbf{55.39} & \textbf{50.39} & \textbf{53.88} \\
        \bottomrule
      \end{tabular}
    } 
    \label{tab:continual_multiple}
  \end{center}
  \vspace{-15pt}
\end{table}

\textbf{Continuous adaptation to a single domain.} We perform continual adaptations for Tent to investigate whether online learning can always improve performance. Since Tent only utilizes AdaBN for normalization, which would degrade performance for some datasets (\eg, VLCS), we also add the comparison to the variants of Tent that update the source statistics online with incoming batch statistics, and we normalize the batch with the updated source statistics. The updating momentum is denoted as $m$. $m=1$ is the original version of Tent and when $m=0$, Tent only utilizes the original source statistics. As shown in \cref{tab:continual}, compared with adaptation only once, the performance of Tent that adapts online can be improved on PACS, VLCS, and OfficeHome with online adaptation when $m>0.5$. However, its performance still cannot surpass our method. Besides, its performance on MiniDomainNet decreases a lot because when the model is not confident about the incoming data, the training on these data may only degrade the performance and online learning would continuously enhance this effect and finally obtain a badly trained model.

\begin{figure}[t]
  \centering
  \includegraphics[width=1.0\linewidth]{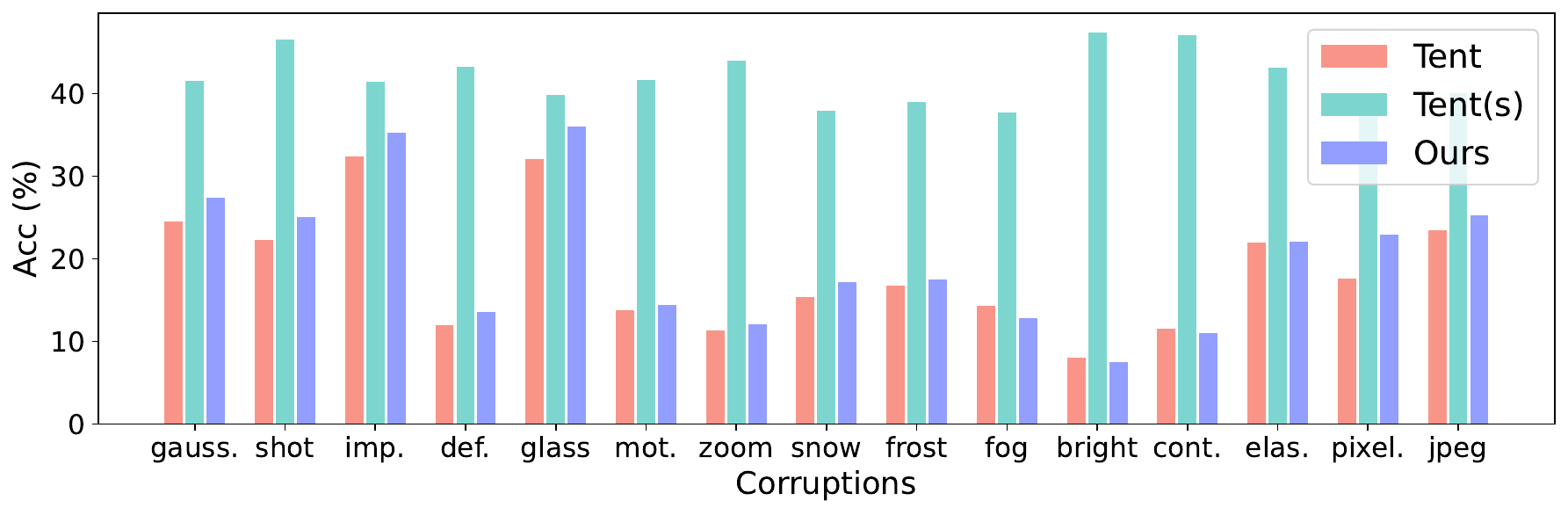}
  \caption{Continuous adaptation on CIFAR-10-C dataset. The trained model is adapted to the batches sampled from shuffled domains. `(s)' means a single domain is adapted.}
  \label{tab:continual_multiple2}
\end{figure}

\label{exp:continous}
\textbf{Continuous adaptation to multiple domains.} Although continuous adaptation to a single domain could improve model performance on some datasets (\eg, PACS), when faced with batches from multiple domains, the performance of Tent drops drastically. We conducted an experiment on the PACS dataset. The result is shown in \cref{tab:continual_multiple}. The baseline is trained on the Photo domain and we adapt it to the other three domains (\ie, Art~(A), Cartoon~(C), Sketch~(S)). When the model is adapted to a single domain, it could improve the performance on both Art and Cartoon significantly (\eg, 67.53\% vs.\ 59.08\% on Art). However, when two domains are incorporated, the improvement drops and it even degrades the performance (\ie, 36.79\% vs.\ 39.90\%) on the environment mixed with Art and Sketch domains. When all three domains are included, there is only little improvement (\ie, 36.33\% vs.\ 35.96\%). Meanwhile, our method could steadily improve the performance of the baseline. To further investigate the effect of multiple domains, we also experiment on CIFAR-10-C dataset. As shown in \cref{tab:continual_multiple2}, when more domains are incorporated, the performance of a multiple-domain adapted Tent drops drastically, while our method could achieve comparable performance to a single-domain adapted Tent.

\begin{figure}[t]
  \centering
  \begin{subfigure}{0.48\linewidth}
    \includegraphics[width=1\linewidth]{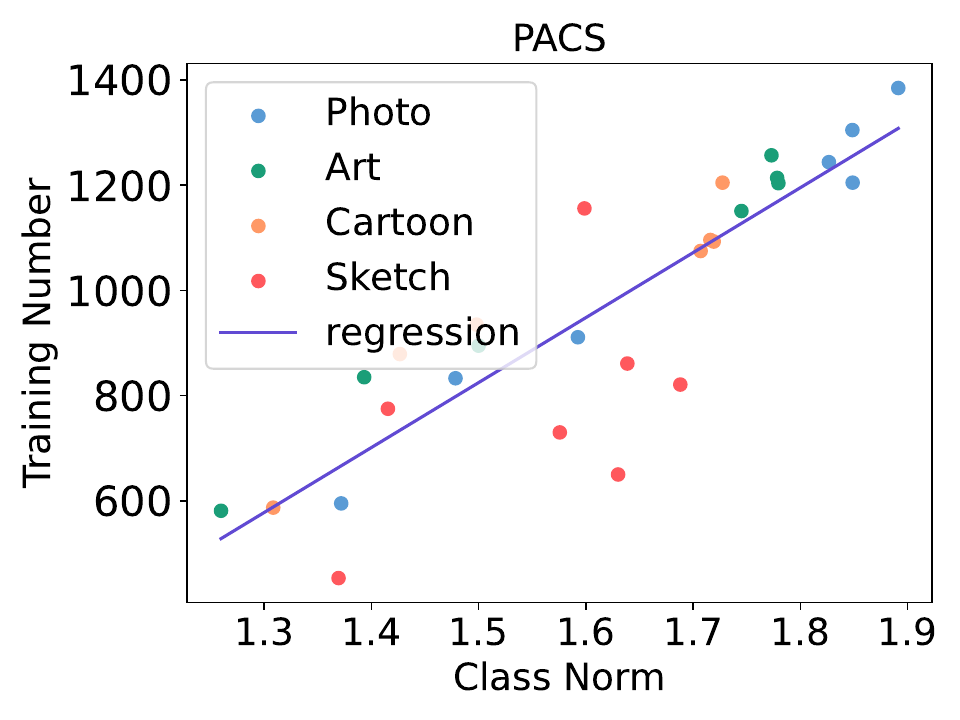} 
  \end{subfigure}
  \begin{subfigure}{0.48\linewidth}
    \includegraphics[width=1\linewidth]{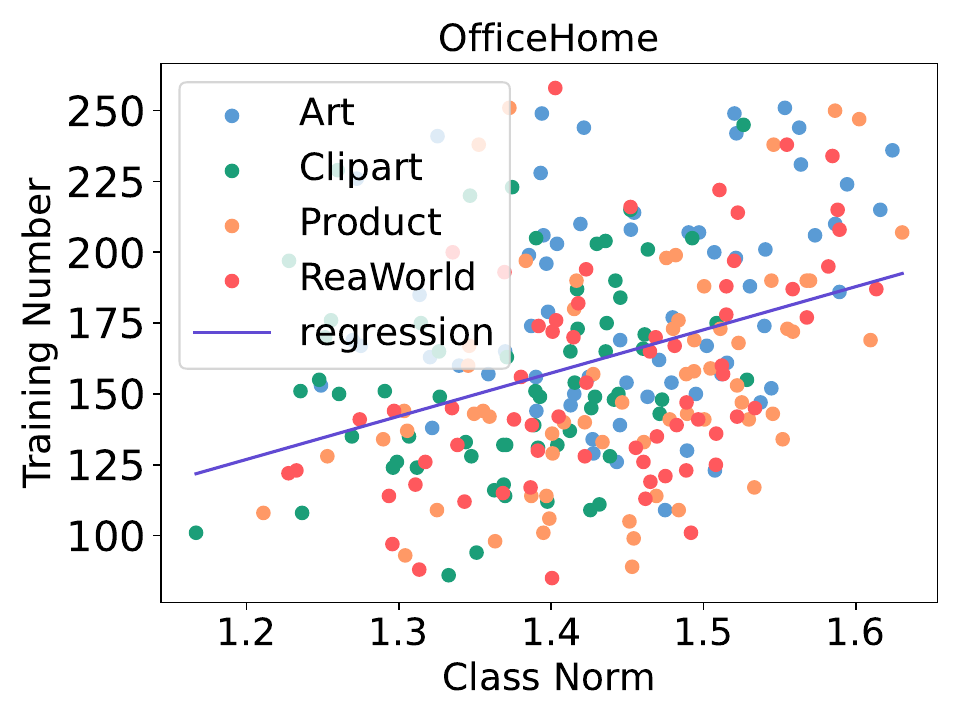}
  \end{subfigure}
  \caption{The relationship between the number of training samples and the norm of classifier weights.}
  \label{fig:classNum}
\end{figure}

\subsection{Further analysis of the norm of classifiers}  
    
% \textbf{The relationship between the number of training samples and the norm of classifier weights.}
In the Experiment section,  we find that our method increases model confidence by changing the angle between the classifier weights and features, which also increases the norm of the feature. With a larger norm, the model could make a more confident prediction. In addition to the norm of the feature, the norm of a classifier also plays an important role in classification. If the weight of a class has a large norm, it would make a more confident prediction. Besides, if the model is trained with a large number of samples for a single class, it would be more confident in this class.
Therefore, we hypothesize that more training samples may have a positive relation to the norm of classifier weights. We plot the correlation between the norm of classifier weights and its corresponding training samples in \cref{fig:classNum}. As seen, there is a positive correlation between these two factors. Note that this phenomenon is more evident on PACS compared to OfficeHome because the training number for each class is relatively small and similar to each other, resulting in a less obvious correlation.

% \subsection{Full results of the comparison to SOTA}

% We provide the full results of the comparison to previous methods in Sec. 4.1 of the main text in \cref{tab_res18,tab_res50}. Note that, we omit the domain names for simplicity.

\end{document}